%% file: ST2Vec.tex
\documentclass[conference]{IEEEtran}
\IEEEoverridecommandlockouts
\usepackage{cite}
\usepackage{amsmath,amssymb,amsfonts}
\usepackage{graphicx}
\usepackage{textcomp}
\usepackage{xcolor}
\usepackage{graphicx}
\usepackage{flushend}
\usepackage{balance}
\usepackage[ruled,vlined]{algorithm2e}
\usepackage{algpseudocode}
\usepackage{multirow}
\usepackage{booktabs}
\usepackage{upgreek}
\usepackage{subfigure}
\usepackage{array}
\usepackage{colortbl}
\usepackage{url}
\usepackage{color}

\newtheorem{definition}{Definition}

\def\BibTeX{{\rm B\kern-.05em{\sc i\kern-.025em b}\kern-.08em
    T\kern-.1667em\lower.7ex\hbox{E}\kern-.125emX}}

\begin{document}

\title{ST2Vec: Spatio-Temporal Trajectory Similarity Learning in Road Networks}
\author{
Ziquan Fang$^{\dagger}$, Yuntao Du$^{\dagger}$, Xinjun Zhu$^{\sharp}$, Danlei Hu$^{\dagger}$, Lu Chen$^{\dagger}$, Yunjun Gao$^{\dagger}$, Christian S. Jensen$^{\mathsection}$\\
\normalsize $^{\dagger}$\emph{College of Computer Science, Zhejiang University, Hangzhou, China} \\
\normalsize $^{\sharp}$\emph{School of Software, Zhejiang University, Ningbo, China}\\
\normalsize $^{\mathsection}$\emph{Department of Computer Science, Aalborg University, Aalborg, Denmark}\\
$^{\dagger}$$^{\sharp}$\emph{\{zqfang, ytdu, xjzhu, dlhu, luchen, gaoyj\}@zju.edu.cn}~~~~\hspace*{0.1in} ~~~~$^{\mathsection}$\emph{\{csj\}@cs.aau.dk} \\
}

\maketitle

\begin{abstract}
  People and vehicle trajectories embody important information of transportation infrastructures, and trajectory similarity computation is functionality in many real-world applications involving trajectory data analysis. Recently, deep-learning based trajectory similarity techniques hold the potential to offer improved efficiency and adaptability over traditional similarity techniques. Nevertheless, the existing trajectory similarity learning proposals emphasize spatial similarity over temporal similarity, making them suboptimal for time-aware analyses.

    To this end, we propose ST2Vec, a trajectory-representation-learning based architecture that considers fine-grained spatial and temporal correlations between pairs of trajectories for spatio-temporal similarity learning in road networks. To the best of our knowledge, this is the first deep-learning proposal for spatio-temporal trajectory similarity analytics. Specifically, ST2Vec encompasses three phases: (i) training data preparation that selects representative training samples; (ii) spatial and temporal modeling that encode spatial and temporal characteristics of trajectories, where a generic temporal modeling module (TMM) is designed; and (iii) spatio-temporal co-attention fusion (STCF), where a unified fusion (UF) approach is developed to help generating unified spatio-temporal trajectory embeddings that capture the spatio-temporal similarity relations between trajectories. Further, inspired by curriculum concept, ST2Vec employs the curriculum learning for model optimization to improve both convergence and effectiveness. An experimental study offers evidence that ST2Vec outperforms all state-of-the-art competitors substantially in terms of effectiveness, efficiency, and scalability, while showing low parameter sensitivity and good model robustness. In addition, two similarity computation based case studies on top-$k$ similarity querying and trajectory clustering offer further insight into the capabilities of ST2Vec.
\end{abstract}


\input{Introduction}

\input{Preliminaries}

\input{Problem}
\input{Framework}

\input{Exp}
\input{Related}

\input{Conclusion}

\bibliographystyle{plain}
\bibliography{ref}

\end{document}

%% file: Introduction.tex
\section{Introduction}
\label{sec:intro}

With the proliferation of GPS-equipped devices and online map based services (e.g., Uber and DiDi), massive volumes of spatio-temporal trajectories of moving objects such as people and vehicles are collected, which motivates various studies of trajectory analytics~\cite{Zheng15, SousaBL20}. A GPS trajectory $T$ is represented as a time-ordered sequence of discrete spatio-temporal points, i.e., $T=\langle(g_1, t_1), (g_2, t_2), ..., (g_n, t_n)\rangle$, where $g$ denotes an observed geo-location and $t$ denotes the corresponding time. A form of trajectory analytics--trajectory similarity computation that evaluates the similarity (distance) between two trajectories benefits a wide range of real-world applications such as ridesharing~\cite{ShangCWJZK18}, traffic analysis~\cite{Zheng15}, social recommendation~\cite{LyeCTHC20} and so on, as depicted in Example 1. 

Example 1. \textit{Given the capability of evaluating the similarity between a pair of trajectories, (i) drivers can be assigned potential ridesharing partners to share ride with; (ii) traffic authorities can predict traffic congestion by aggregating similar trajectories and counting the travel frequencies of roads; and (iii) social apps can identify users with similar living trajectories for friend recommendation. Further, trajectory similarity computation is a fundamental component of downstream similarity-based trajectory analyses, including top-$k$ similarity querying~\cite{shang2017trajectory} and clustering~\cite{YuLCC19}.}

To measure the similarity between two trajectories, a variety of handcrafted distance measures exist, including free space based measures such as DTW~\cite{YiJF98}, LCSS~\cite{vlachos2002discovering}, Hausdorff~\cite{AtevMP10}, and ERP~\cite{ChenN04}, or road network based measures such as TP~\cite{shang2017trajectory}, DITA~\cite{shang2018dita}, LCRS~\cite{Yuan019}, and NetERP~\cite{KoideXI20}. However, these measures are associated with high computation costs. Specifically, they rely on pointwise matching computation~\cite{deeprepresentation}, meaning that they need to scan all point pairs from two trajectories to calculate the similarity scores, which incurs quadratic time complexity $O(\hat{n}^{2})$, where $\hat{n}$ is the average trajectory length. The high computation costs also limit the scalability of a series of downstream similarity-based trajectory analyses. 

To address the above issues, inspired by the success of metric learning in neural language processing~\cite{ChenWZ18, ChenYPCSPQ20} and computer vision~\cite{GeHDS18, Kordopatis-Zilos19}, a new line of studies~\cite{seed, subsimilar, YangW0Q0021, HanWYS021} aims to utilize neural networks to learn trajectory similarities for similarity computation. The core task is to obtain trajectory representations (embeddings) by means of neural networks so that the similarity relations between trajectories are well in that embedding space. This way, the similarity relations between GPS trajectories could be reflected by the similarity relations between the embeddings of the trajectories. Thus, given a pair of trajectories, trajectory similarity learning methods first map trajectories to $d$-dimensional vectors and then calculate the similarities between trajectories based on their embedding vectors, which reduces the time complexity from $O(\hat{n}^{2})$ to $O(d)$, representing a substantial speedup over techniques that operate directly on the GPS trajectories. 

\begin{figure}[t]
	\centering
	\vspace{1mm}
	\includegraphics[width=0.47\textwidth]{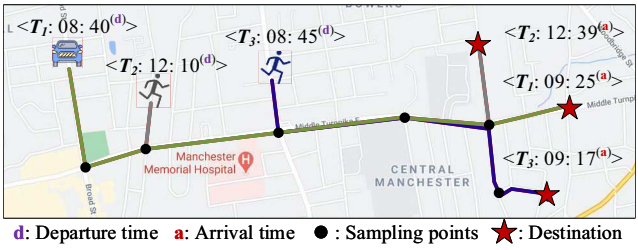}
	\vspace{-4mm}
	\caption{An Illustration of Spatio-Temporal Trajectory Similarity}
	\label{fig:demo}
	\vspace{-5mm}
\end{figure}

While the existing similarity-learning-based trajectory similarity computation approaches~\cite{seed, subsimilar, YangW0Q0021, HanWYS021} successfully improve the high time complexity of traditional similarity computation, they still come with several significant limitations. In particular, all of the above approaches discard the temporal dimensional of spatio-temporal trajectories. That is, they learn and generate spatial-similarity-oriented trajectory embeddings that consider only the spatial dimensional of trajectories, i.e., $T^{(s)}=\langle g_1, g_2, ..., g_n\rangle$. As a result, they can only retrieve spatially similar trajectories, making them inefficient for time-aware scenarios, to be detailed below. 
\noindent
\textbf{Why spatio-temporal similarity?} Unlike existing studies that target only spatially aware trajectory similarity learning and computation, we argue that a general similarity measure should consider both the spatial and the temporal aspects of trajectories. One motivating application is ridesharing. As shown in Fig.~\ref{fig:demo}, $T_1$ denotes the travel planned by a driver, and $T_2$ and $T_3$ belong to two people looking for a ride. The similarities between $T_1$ and $T_2$ and $T_1$ and $T_3$ determine which person to recommend to the driver. Existing spatial-proximity oriented methods typically recommend $T_2$, since $T_1$ and $T_2$ are more spatially close to each other. However, the resulting recommendation is of no use, as $T_1$ and $T_2$ have very different departure times. In spatio-temporal terms, $T_1$ and $T_3$ are most similar, and the person with $T_3$ should get the ride. Overall, taking into account both the spatial and temporal similarity is important in time-aware applications such as transportation planning~\cite{TranMLYHS20} and monitoring~\cite{YuLCZ19}. In addition, time is an essential dimension of spatio-temporal trajectory data and deserves attention on par with the spatial aspect. 

In this paper, we follow an orthogonal but complementary approach to existing space-driven similarity learning studies--we address the problem of spatio-temporal trajectory similarity learning in road networks. To achieve this, a straightforward approach is to cut time into discrete time slots and then perform spatio-temporally similarity computation in each slot using existing spatial similarity learning techniques. However, this approach treats space and time separately and also cannot fully utilize the temporal information due to the coarse-grained discretization of the time dimension. Instead, a more promising direction is to learn unified spatio-temporal embeddings that capture the intricate spatio-temporal similarities between trajectories. Although existing studies~\cite{seed, subsimilar, YangW0Q0021, HanWYS021} offer guidance for spatial embedding, three non-trivial challenges remain to be addressed, including temporal embedding, spatio-temporal fusion, and model optimization.

\textit{Challenge I: How to capture the temporal correlations between trajectories for temporal similarity learning?} The core task is to generate time-oriented embeddings where the temporal similarity relations (i.e., close or distant) between trajectories are preserved. To achieve this, a natural idea is to feed time sequences of trajectories, i.e., $T^{(t)}=\langle t_1, t_2, ..., t_n\rangle$, into recurrent neural network (RNN) models to capture the time sequence information, similarly to how spatial similarity learning that feeds spatial sequences into RNNs. However, temporal modeling is more challenging than spatial modeling. This is because, unlike spatial locations of trajectories are discrete and enable the evaluation of spatial relations by specific measures, the time information exhibits strong continuous and periodic patterns. Specifically, time never stops, resulting in seconds, hours, days, etc. Thus, the time representation must be invariant to time rescaling. Second, trajectories show strong periodicity, which also affects temporal similarity computation. Thus, directly feeding time information into RNNs for temporal dimensional embedding is ineffective since it does not contend with the above problems. Instead, we design a \underline{t}emporal \underline{m}odeling \underline{m}odule, termed TMM, to achieve effective temporal trajectory similarity representation learning. This module is flexible and generic, in that it can be integrated with any existing spatial trajectory similarity learning proposal~\cite{seed, subsimilar, YangW0Q0021, HanWYS021} for spatio-temporal similarity learning.

\textit{Challenge II: How to fuse spatial and temporal trajectory embeddings to achieve unified spatio-temporal similarity learning?} Once the spatial and temporal characteristics are captured, we need to fuse them to generate unified spatio-temporal similarity oriented embeddings. Different users may assign different weights to spatial and temporal similarity, to accommodate applications at hand. For example, applications such as region function estimation~\cite{KongLLTHX19} may assign high importance to spatial aspects of trajectories and thus assign high weight to spatial similarity. In contrast, applications such as ridesharing~\cite{LowalekarVJ21} may assign high importance to the temporal aspects and thus assign high weight to temporal similarity. Overall, a preferable fusion approach must be robust to learn different spatial and temporal weights adaptively and do not hurt model convergence, especially when both the time and spatial dimensions are considered to generate trajectory embeddings. To address this challenge, we develop a \underline{s}patio-\underline{t}emporal \underline{c}o-attention \underline{f}usion module, termed STCF, that fuses the separate spatial and temporal information using a unified fusion approach to obtain unified embeddings.

\textit{Challenge III: How to optimize the models to improve the effectiveness and efficiency?} The two primary goals of learning-based trajectory similarity analyse are effectiveness (similarity querying quality) and efficiency (model convergence speed). Specifically, the training samples, learning procedure, and neural network parameters all potentially affect model performance. To improve effectiveness, we design a new sampling strategy with triplets and then train models using curriculum leaning. To avoid an excess of parameters due to the spatio-temporal modeling and to improve efficiency, we provide two different fusion approaches in the co-attention fusion module.

To address all three challenges, we propose a representation learning based architecture, termed ST2Vec, which leverages fine-grained spatial and temporal information in trajectories to enable unified spatio-temporal similarity learning in road networks. To sum up, we make the following contributions.


\begin{itemize}\setlength{\itemsep}{-\itemsep}
    \item We propose a new representation learning based architecture for spatio-temporal trajectory similarity learning in road networks. To the best of our knowledge, this is the first deep-learning proposal for spatio-temporal similarity computation. ST2Vec is capable of accommodating varying spatial and temporal weights under a series of trajectory measures, thus enabling flexible analyses.

    \item We develop a temporal modeling module for temporal trajectory representation learning. Further, to achieve unified spatio-temporal similarity learning, we develop a spatio-temporal co-attention fusion module with two fusion strategies to integrate the spatial and temporal features of trajectories in an efficient and effective manner. 

    \item For the preparation phase, we improve robustness by developing a new sampling strategy to select representative samples to construct similarity triplets. In the training phase, we exploit the curriculum concept to guide the learning process, further improving the model performance with better accuracy and faster convergence. 

    \item We report on extensive experiments with three real-world data sets and four popular network-aware trajectory measures. The findings offer evidence that ST2Vec is able to outperform four state-of-the-art competitors in terms of effectiveness, efficiency, and scalability. In addition, case studies including top-$k$ similarity querying and clustering demonstrate the downstream capabilities of ST2Vec. 

\end{itemize}

The rest of the paper is organized as follows. Section~\ref{sec:problem} presents preliminaries. Section~\ref{sec:alternative} defines the problem to be solved and explains two alternative approaches to the problem. Section~\ref{sec:method} then details our framework and methods. The experimental results are reported in Section~\ref{sec:exe}. Section~\ref{sec:related} reviews related work. Finally, Section~\ref{sec:conclusion} concludes the paper and offers promising directions.

%% file: Preliminaries.tex
\section{Preliminaries}
\label{sec:problem}

We proceed to introduce key concepts related to the studied problem, including road-network constrained trajectories and the learning targets of ST2Vec.  

\subsection{Road Networks $\&$ Trajectories}
As we target trajectory similarity learning in road networks, we first define road networks and trajectories.
\begin{definition}\label{defn: road}
	{\bf (Road Network)} \textit{A road network is modeled as a directed graph $G = (L, E)$, where $L$ is a set of road vertices and $E \subseteq L \times L$ is an edge set of road segments.}
\end{definition}

Specifically, a vertex $l_i = (x_i, y_i) \in L$ models a road intersection or a road end, in which $x_i$ and $y_i$ denote the longitude and latitude of $l_i$, respectively. An edge $e_{l_i, l_j} \in E$ models a directed road segment from $l_i$ to $l_j$.

The \textbf{GPS trajectory} $T$ of a moving object is initially captured as a time-ordered sequence of sampling points from a GPS device, i.e., $T = \langle (g_1, t_1), (g_2, t_2), ..., (g_n, t_n) \rangle$, where $n$ denotes the length of $T$. Each sampling point is represented as a 2-dimensional (location, time) tuple, i.e., $(g_i, t_i), i \in [1, n]$. Here, $g$ denotes the observed geo-location that consists of longitude and latitude, and $t$ denotes the corresponding time. As we target road-network constrained trajectory similarity learning, we align trajectory points $g$ with vertices $l$ using an existing map-matching procedure (e.g., ~\cite{BrakatsoulasPSW05}). Specifically, we assume the trajectory points are located on the vertices in $G$. It is straightforward to handle trajectory points located on edges: if a point $g$ is located on an edge $e$, we split $e$ into two sub-edges by introducing a new vertex $l_g$. Consequently, each original trajectory $T$ is transformed into a directed path in $G$ from a start vertex to an end vertex, as defined below. 
\begin{definition}\label{defn:trajectory} 
	{\bf (Trajectory)} \textit{Given a road network $G = (L, E)$, a trajectory $T$ is a directed sequence of $m$ $(m \leq n)$ vertices in $G$, i.e., $T = \langle (l_1, t_1), (l_2, t_2), ..., (l_m, t_m) \rangle$, where $l_i \in L$ is a vertex and $t_i$ is the corresponding time.} 
\end{definition}

Unless stated otherwise, we assume in the sequel that trajectories are map matched. Given a trajectory $T$, we use $T^{(s)}$ and $T^{(t)}$ denote its spatial and temporal aspects, respectively, i.e., its \textbf{spatial trajectory} $T^{(s)} = \langle l_1, l_2, ..., l_m \rangle$ and its \textbf{temporal trajectory} $T^{(t)} = \langle t_1, t_2, ..., t_m \rangle$. Note that $T^{(s)}$ and $T^{(t)}$ correspond to each other synchronously at each step. 

\subsection{Spatio-Temporal Similarity $\&$ Learning Targets}
\noindent
\textbf{Remark.} Before performing similarity learning, a similarity measure must be chosen that serves as the learning target. Existing spatial similarity leaning studies~\cite{seed, subsimilar, YangW0Q0021} use free space oriented measures (i.e., Hausdorff~\cite{AtevMP10}, DTW~\cite{YiJF98}, LCSS~\cite{vlachos2002discovering}, and ERP~\cite{ChenN04}) for trajectory similarity learning in free space, or they~\cite{HanWYS021} use network oriented measures (i.e., TP~\cite{shang2017trajectory}) for trajectory similarity learning in road networks. In this paper, without loss of generality, we combine spatial and temporal similarity measures linearly to define spatio-temporal similarity, which is also the learning target of ST2Vec.

Given trajectories $T_i$ and $T_j$, we thus define the spatio-temporal trajectory similarity function $\mathcal{D}(T_i, T_j)$ as a weighted, linear combination of their spatial and temporal similarity. It is simple and flexible to define spatio-temporal similarity this way, and this liner combine approach is popular in previous non-learning-based spatio-temporal trajectory similarity studies~\cite{ShangDZJKZ14, ShangZJYKLW15, shang2017trajectory}. In this paper, we first employ this approach for spatio-temporal trajectory similarity learning. 
\begin{equation}
\label{eq1}
\operatorname{\mathcal{D}}\left(T_{i}, T_{j}\right)=\lambda \cdot \operatorname{\mathcal{D}}_{ S }\left(T_{i}^{(s)}, T_{j}^{(s)}\right)+(1-\lambda) \cdot \operatorname{\mathcal{D}}_{ T }\left(T_{i}^{(t)}, T_{j}^{(t)}\right)
\end{equation} 

Since we study road network constrained trajectory similarity, $\mathcal{D}$ refers to the state-of-the-art network-aware distance measures including TP~\cite{shang2017trajectory}, DITA~\cite{shang2018dita}, LCRS~\cite{Yuan019}, and NetERP~\cite{KoideXI20}. Here, $\mathcal{D}_{S}$ and $\mathcal{D}_{T}$ denote spatial and temporal similarity, respectively. Although these distance measures are predominantly oriented towards spatial proximity, they are also able to support temporal similarity~\cite{shang2017trajectory}. This is because, given a trajectory $T$, its spatial sequence $T^{(s)}$ and temporal sequence $T^{(t)}$ both are time series and support distance aggregation between sequences for similarity evaluations. Since we aim to enable similarity learning across different measures without modifying these measures or their implementations, we do not cover their detailed implementations, but instead refer the interested reader to the literature~\cite{SousaBL20}. Further, parameter $\lambda \in [0, 1]$ controls the relative weight of spatial and temporal similarity, which enables providing flexibility that can be used to support different applications as discussed in Section~\ref{sec:intro}.

%% file: Problem.tex
\section{Problem Statements}
\label{sec:alternative}

We proceed to present the problem formulation, followed by two alternative solutions to our problem. Then, we give a taste of the ST2Vec solution. 

\subsection{Problem Formulation} 
\noindent
\textbf{Problem Statement.} For any pair of trajectories $T_i$ and $T_j$, the spatio-temporal trajectory similarity learning aims to learn a neural-network driven function $\mathcal{G} (\cdot,\cdot)$ such that $\mathcal{G}\left(v_{T_{i}}, v_{T_{j}}\right)$ is maximally close to $\mathcal{D}\left(T_{i}, T_{j}\right)$:
\begin{equation}
\label{eq2}
\arg \min \limits_{\mathcal{\mathcal{M}}} \left|\mathcal{G}\left(v_{T_{i}}, v_{T_{j}}\right)-\mathcal{D}\left(T_{i}, T_{j}\right)\right|
\end{equation}

Here, $\mathcal{M}$ denotes the model parameters of the neural network, $\mathcal{D}$ is the spatio-temporal trajectory similarity defined in Eq.~\ref{eq1}, and $v_{T_i}$ and $v_{T_j}$ are the spatio-temporal embeddings of $T_i$ and $T_j$. According to Eq.~\ref{eq2}, spatio-temporal similarity learning aims to train a neural network that realizes a function $\mathcal{G}$ by embedding trajectories (i.e., $T_i$ and $T_j$) into low-dimensional vectors (i.e., $v_{T_i}$ and $v_{T_j}$) that reflect their similarity relations. That is, $v_{T_i}$ and $v_{T_j}$ are close (resp. distant) to each other if $T_i$ and $T_j$ are similar (resp. dissimilar) to each other. 

\subsection{Alternative Solutions} To realize spatio-temporal trajectory similarity computation, two alternative solutions exist that are extensions of existing spatial similarity learning proposals~\cite{seed, subsimilar, YangW0Q0021, HanWYS021}. 

A straightforward solution is to split the time axis into discrete time intervals and then assign trajectories to different time intervals using sliding-window methods. After this pre-processing, one can conduct similarity computation in each time interval using existing spatial similarity learning methods. However, this approach is spatially-oriented, is coarse-grained, and is suboptimal. Further, time is continuous and unbounded, making it difficult to determine an appropriate window length, and regardless of the length chosen, inaccurate or incorrect spatio-temporal similarity computations are inevitable. In addition, this approach relies on discrete time and trajectory processing, which incurs additional processing costs. 

Another solution to capture the temporal information of trajectories is to feed the time sequences of trajectories to RNNs the same way that location sequences of trajectories are fed to RNNs. Then, the resulting temporal vectors can be combined with the spatial vectors obtained by existing methods~\cite{seed, subsimilar, YangW0Q0021, HanWYS021} to achieve spatio-temporal similarity learning. Although this approach is more reasonable than the first, it is also naive. As discussed in Section~\ref{sec:intro}, temporal correlations are more complex than spatial correlations because time is continuous and correlations may be periodic. Consequently, simply applying spatial trajectory embedding methods to embed time is likely to be sub-optimal. The paper's experimental study considers the above two approaches and provides detailed insight into their performance. 

\subsection{ST2Vec Solution}
In contrast to above solutions, we propose a new \textbf{representation learning} based architecture, termed ST2Vec, that is capable of exploiting the fine-grained temporal and spatial information in trajectories to enable unified spatio-temporal similarity learning. Fig.~\ref{fig:architecture} shows the architecture and training scheme of ST2Vec. It takes similar and dissimilar pairs of anchor trajectories to construct input \textbf{similarity triplets} that consider both the spatial (i.e., $T^{(s)}$) and temporal (i.e., $T^{(t)}$) dimensions. Then, ST2Vec learns to embed trajectories, mapping the trajectories to low-dimensional space, which process is shown in the dashed rectangle in Fig.~\ref{fig:architecture}. This process proceeds until the trajectory similarities evaluated on the embedding vectors (denoted by light blue and light yellow cubes) approximate the ground-truth similarities (denoted by blue and yellow cubes) as computed by Eq.~\ref{eq1}. 

In order to generate the spatio-temporal similarity-oriented embeddings (cf.\ the green rectangle with reddish edges in Fig.~\ref{fig:architecture}), we must capture the temporal and spatial information in trajectories and fuse this information in a unified manner. To achieve this, STVec features three major modules, i.e., \underline{t}emporal \underline{m}odeling \underline{m}odule (TMM), \underline{s}patial  \underline{m}odeling \underline{m}odule (SMM), and \underline{s}patial-\underline{t}emporal \underline{c}o-attention \underline{f}usion module (STCF). These modules are covered next. We note that this design with three modules makes it possible to replace our SMM with any existing spatial similarity learning module to realize spatio-temporally aware similarity learning.


%% file: Framework.tex
\section{The ST2Vec Approach}
\label{sec:method}

We first detail the three modules. Then, we describe the training process of representation-based trajectory similarity learning. Finally, we provide an analysis of ST2Vec approach. 

\begin{figure}[t]
	\centering
	\hspace{-2mm}
	\includegraphics[width=0.5\textwidth]{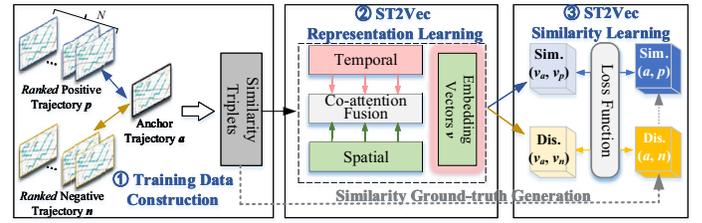}
	\vspace{-4mm}
	\caption{Architecture and Training Scheme of ST2Vec}
	\label{fig:architecture}
	\vspace{-4mm}
\end{figure}

\begin{figure*}[tb]
	\centering
	\vspace{-3mm}
	\includegraphics[width=0.98\textwidth]{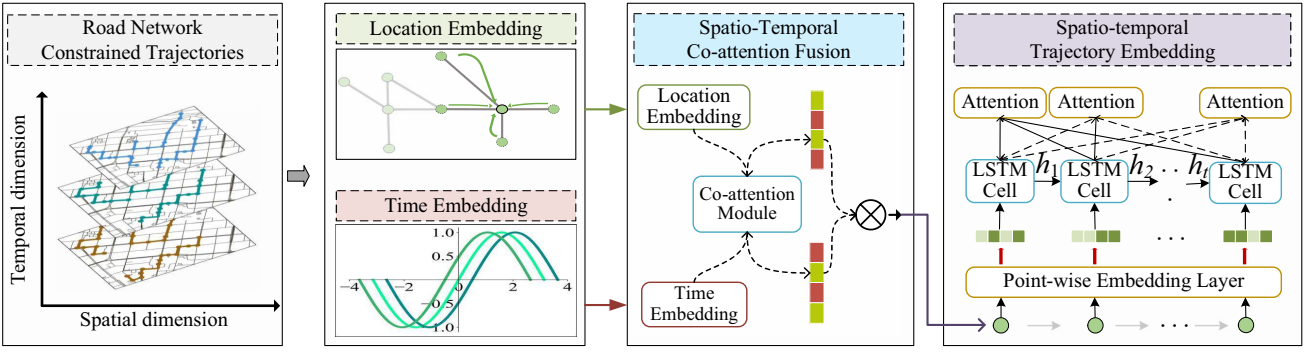}
	\vspace{-3mm}
	\caption{An Overview of the ST2Vec with \textit{Unified Embedding}}
	\label{fig:overview}
	\vspace{-4mm}
\end{figure*}

\subsection{Temporal Modeling Module (TMM)} 
To capture the correlations between a pair of temporal trajectories ($T_i^{(t)}$, $T_j^{(t)}$), it is natural to use state-of-the-art sequence models such as RNN, LSTM, or their variants, to embed temporal trajectories into vectors. However, this fails to handle time's periodic and non-periodic temporal patterns. 

\noindent
\textbf{Basic idea.} To achieve fine-grained temporal representation learning, we integrate \textit{time embedding} with \textit{temporal sequence embedding} to construct a trajectory-aware temporal sequence modeling module. Further, we notice that different time points may have different importance, e.g., rush hour vs. late night. Thus we further introduce the attention function to enhance the representation of temporal irregularity. 

\subsubsection{\textbf{Time Embedding}} Inspired by position embedding in BERT~\cite{WangSLJYLS21}, for each time point $t$ in a temporal trajectory, we learn its time embedding $t'$, which is a vector of size $q + 1$. 
\begin{equation}
\label{eq:time2vec}
t'[i]= \begin{cases}\omega_{i} t+\varphi_{i}, & \text { if } i=0 \\ \cos \left(\omega_{i} t+\varphi_{i}\right), & \text { if } 1 \leq i \leq q\end{cases}
\end{equation}

Here, $t'[i]$ denotes the $i$-th element of $t'$, $\omega_0, ..., \omega_l$ and $\varphi_0, ..., \varphi_l$ are learnable parameters, and $\cos (\cdot,\cdot)$ serves as a periodic activation function that helps capture periodic behaviors without the need for feature engineering. For $1 \leq i \leq q$, $\omega_i$ and $\varphi_i$ are the frequency and the phase-shift of the cos function, and thus the period of the cos function is $\frac{2\pi}{\omega_i}$, i.e., it has the same value at $t$ and $t+\frac{2\pi}{\omega_i}$. The linear term represents the progression of time and can be used for capturing non-periodic patterns in the input that depend on time. Based on this, we can embed a temporal trajectory $T^{(t)}$ into a sequence of time vectors, i.e., $\langle t_1, t_2, ..., t_m \rangle \rightarrow \langle t_1', t_2', ..., t_m' \rangle$. 

\subsubsection{\textbf{Temporal Sequence Embedding}} As illustrated in Fig.~\ref{fig:overview}, if we remove the spatio-temporal co-attention fusion module, after embedding each time point in a trajectory, we could feed $\langle t_1', t_2', ..., t_m' \rangle$ to an LSTM to model its temporal dependence. The recurrent step of an LSTM is performed as follows. At each step $i$, an LSTM cell takes as input the current input vector $x_i$ and the state of the previous step $h_{i-1}$, and it outputs the state vector of the current step $h_i$. 
\begin{equation}
h _{i}=\operatorname{LSTM}\left(t _{i}^\prime, h _{i-1}, i_i, f_i, o_i, m_i\right),
\end{equation}
\noindent
where $i_i$, $f_i$, $o_i$, and $s_i$ represent an input gate, a forget gate, an output gate, and a memory cell, respectively. More details on LSTMs are available elsewhere~\cite{BreuerEJBHF19}. In the context of our LSTM layer, $t_i^\prime$ is the learned time embedding that corresponds to the original time $t_i$. The LSTM unit exploits the embedded time, the hidden state, and the cell state from the previous step to compute the new hidden state and to update the cell state. Eventually, we treat the last hidden state $h_{t}$ as the deep temporal trajectory representation because it contains all temporal information of the trajectory. Overall, a temporal information preserving representation is learned by the recurrent procedure that processes the time points and captures the correlations among them. 

\subsubsection{\textbf{Decoupled Attention}} Different time points in a trajectory have different weights in computations. To contend with this, we employ attention mechanisms to capture the correlations between trajectory points to improve model effectiveness, to be verified experimentally. Specifically, we propose a self-attention mechanism to compute the attention score between time points in the same trajectory as follows.
\begin{equation}
\tilde{ h }_{i}^{(p)}=\sum_{k=1}^{i} \operatorname{att}\left( h _{i}^{(p)}, h _{k}^{(p)}\right) \cdot h _{k}^{(p)}
\end{equation} 

Here, $\tilde{ h }_{i}^{(p)}$ denotes the improved state representation, and att$(\cdot,\cdot)$ is an attention function:
\begin{equation}
\operatorname{att}\left( h _{i}^{(p)}, h _{k}^{(p)}\right)=\frac{\alpha_{i, k}}{\sum_{k^{\prime}=1}^{i} \exp \left(\alpha_{i, k^{\prime}}\right)}
\end{equation} 
where, $\alpha_{i, k}= w _{1}^{\top} \cdot \tanh \left( W _{1} \cdot h _{k}^{(p)}+ W _{2} \cdot h _{ i }^{(p)}\right)$ and $w_1$, $W_1$, and $W_2$ are the parameter vector and matrices to learn. By including the attention mechanism into the temporal sequence embedding, we can discover more important time points, in turn improving model performance, to be confirmed experimentally. Note that we also use the hidden representation of the last step to encode the full temporal trajectory. 

\subsection{Spatial Modeling Module (SMM)}
\noindent
\textbf{SMM vs. Previous Studies.} Since several studies exist on spatial trajectory modeling, we first detail the main difference between them and ST2Vec. Most of the previous studies~\cite{seed, subsimilar, YangW0Q0021} measure trajectory similarities in free space. In this setting, RNN-type models are adopted widely to capture the sequence information for spatial similarity representation learning. However, moving objects such as people and vehicles move in road networks~\cite{ShangCWJZK18}, in which case, these studies do not reflect the real distances between trajectories due to the movement restrictions imposed by road networks. Further, such restrictions cannot be learned by single RNN models. To this end, the state-of-the-art study~\cite{HanWYS021} combines GNNs with LSTM for road network constrained trajectory representation learning and it achieves the state-of-the-art similarity learning performance. However, it is designed specifically for POI (Points of Interest) based similarity computation. That is, the study~\cite{HanWYS021} treats two trajectories $T_i$ and $T_j$ as similar if they share the same POIs. This approach gives more significance to POIs while ignoring detailed travel paths, which might yield inaccuracies when measuring the similarity between trajectories that share the same POIs but have different moving paths. In contrast to all of the above studies, we target fine-grained spatial similarity learning in road networks, which considers both the locations (i.e., sampling points) and paths when evaluating the similarity between two trajectories. 

\noindent
\textbf{Basic idea.} Given a spatial trajectory $T^{(s)} = \langle l_1, l_2, ..., l_m \rangle$ ($l_i$ denote vertices in $G$), we aim to embed $T^{(s)}$ as a vector $v_{T^{(s)}}$ in low-dimensional space that captures to capture its road-network constrained spatial information. Due to the spatial dependencies in the underlying road network, it is naturally to utilize GNNs to take into account the structure of $G$, as GNNs have been used successfully in road-network settings like region classification~\cite{YangWWCW19} and traffic prediction~\cite{ZhengFW020}. Hence, to achieve spatial similarity oriented representation learning, we develop spatial modeling module (SMM), which also encompasses three phases, i.e., location embedding, spatial sequence embedding, and spatial attention. 

\subsubsection{\textbf{Location Embedding}} Trajectories of objects (e.g., people and vehicles) moving in road networks are constrained by the topology of the road network. Thus, the distance between two spatially close sampling points can still be large, if the points are not connected well in the road network. To capture  topological, or structural, information, we first utilize the Node2Vec~\cite{node2vec-kdd2016} method, which aims to capture the co-occurrence of the adjacent locations in road networks. Specifically, given a vertex $l_i$, we adopt Node2Vec to approximate the spatial conditional probability of vertices in its neighborhood, i.e., we perform the mapping $l_i \rightarrow n_i$, where $l_i$ and $n_i$ denote the original and embedded locations, respectively. Then, locations sharing similar neighborhoods tend to have similar embeddings. Next, we feed the embedded locations (i.e., the $n_i$) to a GNN step by step to obtain locally smoothed location embeddings, where spatially adjacent locations tend to be close in the latent space. Given a road network $G$ and a low-dimensional representation $n_i$ of location $l_i \in G$, we define the GCN function as follows.
\begin{equation}
\label{eq:GCN}
l_{i}^\prime=\operatorname{GCN}\left( n_i\right)=\sigma\left(\left(\sum_{j \in N _{i}} c_{i j} W _{s} n_j\right) \| n_i\right)
\end{equation}

Here, $l_i^\prime$ is a vertex/location representation, $\sigma$ is a non-linear activation function, $c_{ij}$ is an adjacency weight, $W_{s} \in \mathcal{R}^{d \times d}$ is learnable matrix shared by all vertices in $G$, $||$ denotes the concatenation operation, and $\mathcal{N}_i$ is the set of neighbor vertices of $n_i$ in $G$. Based on Node2Vec and Eq.~\ref{eq:GCN}, we obtain a fine-grained representation of each spatial trajectory, i.e., $\langle l_1, l_2, ..., l_m \rangle \rightarrow \langle l_1^\prime, l_2^\prime, ..., l_m^\prime \rangle$.

\subsubsection{\textbf{Spatial Sequence Embedding $\&$ Attention}} As illustrated in Fig.~\ref{fig:overview}, if we remove the spatio-temporal co-attention fusion module, we can obtain a sequence of location vectors as input for the LSTM model. The spatial sequence embedding here is similar to the temporal sequence embedding in Eq.~4. 

Given a spatial trajectory $T^{(s)}$, based on Node2Vec and Eq.~\ref{eq:GCN}, we first obtain its initialized location sequence representation and feed that to a LSTM model to encode the spatial information. Further, a self-attention mechanism is applied to capture different contributions of the different locations in the learning process. We do this because different location points in a trajectory contribute differently to the similarity computation. For instance, noisy location points with obvious deviations from other points typically have high influence on the similarity computation. Finally, we use the hidden state of the last step of the LSTM model as the spatial embedding. 

\subsection{Spatio-Temporal Co-attention Fusion (STCF)}
Next, we propose to fuse the hidden spatial and temporal information of trajectories to generate spatio-temporal oriented embeddings. We propose a spatio-temporal co-attention fusion module that uses two fusion strategies. 

\subsubsection{Separate Fusion (SF)} Based on \textit{temporal sequence embedding} (Section~\ref{sec:method}-A) and \textit{spatial sequence embedding} (Section~\ref{sec:method}-B), we could embed temporal and spatial trajectories separately. Thus, a straightforward approach is to first generate spatial and temporal embeddings of trajectories with two separate LSTM models and then combine the two types of embeddings. Given a trajectory $T$ with its initial temporal embedding $(t_1^\prime, t_2^\prime, ..., t_m^\prime)$ and initial spatial embedding $(l_1^\prime, l_2^\prime, ..., l_m^\prime)$, we define the spatio-temporal trajectory embedding based on the separate fusion as:
\begin{equation}
v_T =\textit{LSTM}_t(t_1^\prime, t_2^\prime, ..., t_m^\prime) + \textit{LSTM}_s(l_1^\prime, l_2^\prime, ..., l_m^\prime)
\end{equation}

Although this approach is simple and effective, it requires two LSTM models to separately capture the temporal information and spatial information, doubling the number of parameters that need to be determined in LSTMs. To improve model convergence/efficiency, we propose another fusion strategy.

\subsubsection{Unified Fusion (SF)} Given a trajectory, based on the aforementioned procedure of \textit{time embedding} and \textit{location embedding}, we could obtain its initial temporal sequence embedding, denoted by $\tau^{(t)} = \langle t_1^\prime, t_2^\prime, ..., t_m^\prime \rangle$, and its initial spatial sequence embedding, denoted by $\tau^{(s)} = \langle l_1^\prime, l_2^\prime, ..., l_m^\prime \rangle$. Since these representations capture different dimensions of trajectory properties, we design a co-attention fusion module to enhance them by letting them interact with each other, as depicted in Fig.~\ref{fig:overview}. Specifically, we first make a transformation for the temporal and spatial features via a matrix $W_F$.
\begin{equation}
z_{\tau}^1=W_{F} {\tau}^{(t)}, \quad z_{\tau}^2=W_{F} {\tau}^{(s)}
\end{equation}
The interaction between two representations is calculated by
\begin{equation}
\begin{gathered}
\beta_{i, j}=\frac{\exp \left(W_{Q^{\prime}} z_{\tau}^{i} \cdot W_{K}^{\prime} z_{\tau}^{j^{T}}\right)}{\sum_{j^{\prime} \in\{1,2\}} \exp \left(W_{Q^{\prime}} z_{\tau}^{i} \cdot W_{K}^{\prime} z_{\tau}^{j^{\prime} T}\right)}, \\
{\tau}^{\hat{(t)}}=\operatorname{Norm}\left(\textit{FFN}^{\prime}\left(\beta_{1,1} z_{\tau}^{1}+\beta_{1,2} z_{\tau}^{2}\right)+\tau^{(t)}\right), \\
{\tau}^{\hat{(s)}}=\operatorname{Norm}\left(\textit{FFN}^{\prime}\left(\beta_{2,1} z_{\tau}^{1}+\beta_{2,2} z_{\tau}^{2}\right)+\tau^{(s)}\right),
\end{gathered}
\end{equation}
Here, $W_Q^{\prime}$ and $W_K^{\prime}$ are matrices with the same shape as $W_F$, and ${\tau}^{\hat{(t)}}$ and ${\tau}^{\hat{(s)}}$ are the enhanced representations of ${\tau}^{(t)}$ and ${\tau}^{(s)}$. As shown in Fig~\ref{fig:overview}, we then feed the enhanced initial temporal and spatial sequence embeddings into the same, single LSTM architecture for unified spatio-temporal trajectory embedding. This type of fusion manner is formally defined as follows. 
\begin{equation}
v_T = \textit{LSTM}({\tau}^{\hat{(t)}}, {\tau}^{\hat{(s)}})
\end{equation}

\subsection{Training and Model Optimization}\label{data}

\subsubsection{Training Data Selection} Training sample selection is essential to the similarity learning~\cite{FaghriFKF18}. Recall that we aim to minimize the difference between the learned similarity $\mathcal{G}(v_{T_i}, v_{T_j})$ and the ground truth similarity $\mathcal{D}(T_i, T_j)$, where $\mathcal{G}$ denotes the target neural network and $\mathcal{D}$ represents some chosen distance measure. Hence, training samples $(T_i, T_j)$ are required. Guided by the similarities generated from training samples, ST2Vec trains a neural network to yield embeddings that approximate the chosen similarity function as defined in Eq.~1. A simple approach is to use all pairs of trajectories as training samples, but this incurs excessive training costs and causes overfitting. Thus, given a trajectory dataset, how to select samples to supervise the training process is important. 

\noindent
\textbf{Selection Strategy.} Given a trajectory dataset, we randomly select one trajectory as an anchor $T_a$ and sample a similar (resp. dissimilar) trajectory as its positive $T_p$ (resp. its negative $T_n$) trajectory. Such a triple of an anchor, a positive, and a negative trajectory form a similarity triplet $(T_a, T_p, T_n)$. The triplets provide trajectory samples in terms of similarities and dissimilarities, making the trained model effective and robust. This type of sampling is used widely in imagine classification~\cite{WangWZJGZL018} and text clustering~\cite{XieGF16}. Specifically, as depicted in Fig.~\ref{fig:architecture}, when we select an anchor trajectory, we find its $N$ most similar trajectories as \textbf{similar ones}. Disregarding the similar trajectories, we randomly select $N$ other trajectories as \textbf{dissimilar ones}, such a sampling strategy provides a trade off between robustness and efficiency.  



\subsubsection{Training Process} In the training data selection phase, we obtain representative similarity triplets $(T_a, T_p, T_n)$. In the modeling phase, ST2Vec embeds the triples of trajectories considering both temporal and spatial aspects, i.e., $f_\theta^{(t, s)}: (T_a^{(t, s)}, T_p^{(t, s)}, T_n^{(t, s)}) \rightarrow (v_a^{(t, s)}, v_p^{(t, s)}, v_n^{(t, s)})$, where superscripts $t$ and $s$ denote the temporal and spatial aspects, respectively, and $f_\theta^{(t, s)}$ provides the functionality of $\mathcal{G}$. Also, the spatio-temporal similarities can be computed by the $L_2$ norm based on $\|v_{a}-v_{p}\|_{2}$ and $\|v_{a}-v_{n}\|_{2}$. The ground truth similarity $\mathcal{D}_{{a}, {p}}$ can be normalized as $\mathcal{D}_{a, p}^{\prime}=\exp (-\alpha \cdot \mathcal{D}_{a, p}) \in[0,1]$, and the dissimilarity $\mathcal{D}_{{a}, {n}}$ can be normalized as $\mathcal{D}_{a, n}^{\prime}=\exp (-\alpha \cdot \mathcal{D}_{a, n}) \in[0,1]$. Note $\alpha$ is a tunable parameter, making it possible to control the scale of similarity values. 

\noindent
\textbf{Loss Function.} We define a space and time aware loss function $\mathcal{L}$ that measures the weighted sum squared errors of similarity triplets.
\begin{equation}
\begin{aligned}
\mathcal{L} = & \mathcal{D}_{a, p}^{\prime}\left(\mathcal{D}_{a, p}^{\prime}-\exp \left(-\left\|v_{a}-v_{p}\right\|_{2}\right)\right)^{2} \\
&+\mathcal{D}_{a, n}^{\prime}\left(\mathcal{D}_{a, n}^{\prime}-\exp \left(-\left\|v_{a}-v_{n}\right\|_{2}\right)\right)^{2}
\end{aligned}
\end{equation}
As before, subscripts $a$, $p$, and $n$ indicate anchor, positive, and negative, respectively, and $\mathcal{D}$ is the spatio-temporal similarity function defined in Eq.~1. It is worth mentioning that, this design combines the spatial similarity and the temporal similarity into a unified measure, which enables ST2Vec to adapt varying spatial and temporal weights according to different preferences, regardless of what $\lambda$ is.

\subsubsection{Training Optimization} We observe that the existing trajectory similarity learning methods typically use random training instances for learning and often converge slowly. Recently studies in text generation~\cite{abs-2102-03554}, translation~\cite{LiuLWC20}, and object detection~\cite{SovianyIRS21} suggest that using training samples from easy to hard, i.e., first training easy ones and then hard ones, benefits the learning process. Such an organization of learning in human learning is referred to as a curriculum learning. In view of this, given a trajectory anchor $T_a$ and its $k$ similar (dissimilar) ones, we can get $k$ training triplets. As shown in Fig.~\ref{fig:architecture}, we can order the triplets with the easy ones first (i.e., the most dissimilar to $T_a$), followed by the hard ones (i.e., the most similar to $T_a$). Then, we feed those triplets from the easy to hard. This way, ST2Vec achieves faster convergence and higher accuracy, to be validated experimentally. 

\begin{table*}[tb]
\vspace{-5mm}
\caption{The Comparison of Similarity Learning on TP, DITA, LCRS, and NetERP Distances using T-Drive Dataset}
\vspace{-2.5mm}
\hspace{-5mm}
\begin{tabular}{p{1.4cm}<{\centering}|p{1.6cm}<{\centering}|p{0.82cm}<{\centering}p{0.82cm}<{\centering}p{0.86cm}<{\centering}|p{0.82cm}<{\centering}p{0.82cm}<{\centering}p{0.86cm}<{\centering}|p{0.82cm}<{\centering}p{0.82cm}<{\centering}p{0.86cm}<{\centering}|p{0.82cm}<{\centering}p{0.82cm}<{\centering}p{0.86cm}<{\centering}}
\hline
\multirow{2}{*}{Category} & \multirow{2}{*}{Methods} & \multicolumn{3}{c|}{TP [22]}            & \multicolumn{3}{c|}{DITA [26]}                & \multicolumn{3}{c|}{LCRS [44]}   & \multicolumn{3}{c}{NetERP [14]}  \\ \cline{3-14}
                          &                          & HR@10          & HR@50          & R10@50         & HR@10          & HR@50          & R10@50         & HR@10      & HR@50      & R10@50     & HR@10      & HR@50      & R10@50     \\ \hline
\multirow{4}{*}{\begin{tabular}[c]{@{}c@{}}Window\\ Guided \\ Baselines\end{tabular}}  
& NEUTRAJ$^\textit{w}$                    & 0.0978	&0.1373	&0.1582	&0.0805	&0.1243	&0.1442	&0.0357	&0.0419	&0.0861	&0.0054	&0.0173	&0.0198  \\
& Traj2SimVec$^\textit{w}$                    & 0.0827	&0.1261	&0.1397	&0.053	&0.0682	&0.1151	&0.016	&0.098	&0.1861	&0.0209	&0.0986	&0.1010  \\
& T3S$^\textit{w}$                    &0.1295	&0.1733	&0.2045	&0.0838	&0.1266	&0.1489	&0.0435	&0.0678	&0.1187	&0.01253	&0.0292	&0.0388 \\
& GTS$^\textit{w}$                    & 0.3034	&0.3980	&0.6975	&0.1178	&0.2223	&0.3991	&0.0188	&0.0538	&0.0652	&0.0252	&0.0408	&0.0505  \\ \hline
\multirow{4}{*}{\begin{tabular}[c]{@{}c@{}}LSTM\\ Guided \\ Baselines\end{tabular}}  
& NEUTRAJ$^\textit{l}$                    & 0.1765	&0.2221	&0.2703	&0.0767	&0.1103	&0.1340	&0.0533	&0.1126	&0.1694	&0.0259	&0.0502	&0.0736  \\
& Traj2SimVec$^\textit{l}$                    & 0.1446	&0.1902	&0.2263	&0.05261	&0.0642	&0.1071	&0.0329	&0.1397	&0.2257	&0.0328	&0.1050	&0.1244  \\
& T3S$^\textit{l}$                    & 0.1535	&0.1984	&0.2382	&0.0806	&0.1191	&0.1422	&0.0486	&0.0904	&0.1445	&0.0193	&0.0398	&0.0563  \\
& GTS$^\textit{l}$                    &0.3709	&0.4756	&0.7965	&0.1277	&0.2321	&0.4143	&0.0360	&0.1074	&0.1342	&0.03984	&0.0655	&0.0894  \\ \hline
\multirow{4}{*}{\begin{tabular}[c]{@{}c@{}}Our TMM\\ Guided \\ Baselines\end{tabular}}  
& NEUTRAJ$^\textit{t}$                    & 0.3371          & 0.4091          & 0.7001          & 0.1412          & 0.2719          & 0.4892         & 0.0924          & 0.2848         & 0.3632          & 0.1086          & 0.1832          & 0.2841  \\
& Traj2SimVec$^\textit{t}$                    & 0.3987          & 0.5364          & 0.6593        & 0.1321         & 0.3072          & 0.3643        & 0.0968          & 0.2826          & 0.3741         & 0.2128          & 0.3212          & 0.5553  \\
& T3S$^\textit{t}$                    & 0.3944          & 0.5011          & 0.7917          & 0.1284          & 0.2288          & 0.4073          & 0.1442          & 0.4331          & 0.5672          & 0.1464          & 0.2767          & 0.4077  \\
& GTS$^\textit{t}$                    & 0.4243          & 0.5640          & 0.8026          & 0.3244          & 0.4370          & 0.6381          & 0.1643          & 0.4427          & 0.6242          & 0.2154          & 0.3477          & 0.5343  \\ \hline
\multirow{1}{*}{Our Methods}  
& ST2Vec                    & \textbf{0.4624}         & \textbf{0.5868}         & \textbf{0.8361}          & \textbf{0.3773}         & \textbf{0.5037}          & \textbf{0.7031}          &\textbf{0.1806}         & \textbf{0.5469}          & \textbf{0.7293}       & \textbf{0.2386}          & \textbf{0.3493}       & \textbf{0.6133} \\ \hline
\end{tabular}
\label{tab:comparisonTdrive}
\vspace{0mm}
\end{table*}

\subsection{Approach Analysis} Let's go back to the essence of the similarity learning, which aims to reduce the time complexity of traditional measures using pair-wise computations on original GPS trajectories, by performing similarity computation based on embedding vectors. To compute the similarity between a pair of trajectories in a road network, the time complexity of pair-wise based methods is $O ((E + L\lg L) \cdot m^{2})$, where $O(E + L\lg L)$ is the cost of finding a shortest path between two vertices and $m$ is the average trajectory length. Consequently, traditional methods cannot be applied efficiently in downstream tasks such as clustering, where the distances between all trajectory pairs must be computed. In contrast, the time complexity of ST2Vec for trajectory similarity computation is $O(d)$, where $d$ is a constant dimension. Thus, ST2Vec is more efficient for large-scale trajectory data analysis, as verified in Table~\ref{table:time}. Once $\mathcal{G}$ is well-trained, it enable computing the inter-trajectory spatio-temporal similarity in linear time, since $v_{T_i}$ and $v_{T_j}$ are low-dimensional vectors.

%% file: Exp.tex
\section{Experimental Study}
\label{sec:exe}
We first describe the experimental settings and then compare the effectiveness of ST2Vec with popular and state-of-art baselines. Next, we evaluate model efficiency and scalability. Further, we provide detailed insight into parameter sensitivity to characterize the robustness of ST2Vec. In addition, we include ablation analyses. Moreover, we report on the acceleration capability of ST2Vec over traditional non-learning based measures. Last but not least, we perform two case studies to examine ST2Vec intuitively. 

\subsection{Experimental Settings}
\noindent
\textbf{Datasets.} In the experiments, three public real-life trajectory data sets are adopted for experimental evaluations, including T-Drive$\footnote{\footnotesize https://www.microsoft.com/en-us/research/publication/t-drive-trajectory-data-sample/}$, Rome$\footnote{\footnotesize https://crawdad.org/roma/taxi/20140717/}$, and Xi'an$\footnote{\footnotesize https://outreach.didichuxing.com/research/opendata/}$.

\begin{itemize}\setlength{\itemsep}{-\itemsep}
\item \textbf{T-Drive} contains 15 million taxi trajectory points from Beijing, China, collected from Feb. 2 to Feb. 8, 2008.
\item \textbf{Rome} includes 367,052 trajectories from taxis in Rome, Italy, covering 30+ days.
\item \textbf{Xi'an} contains 806,482 trajectories from Xi'an, China, collected during one weak by the DiDi company.
\end{itemize}

Since we target trajectory similarity analytics in road networks, we map match~\cite{BrakatsoulasPSW05} all trajectories to the corresponding road networks from OpenStreetMap. This way, the raw GPS trajectory data is transformed into time-ordered vertex sequences, in accordance with Definition~\ref{defn:trajectory}. Further, we acquire trajectories from urban areas and remove trajectories with fewer than 10 sampling points. This preporcessing yields 348,210 trajectories in T-Drive, 45,157 trajectories in Rome, and 553,016 trajectories in Xi'an. 

\noindent
\textbf{Evaluation Metrics and Ground-truth.} Following existing trajectory similarity leaning studies~\cite{seed, subsimilar, YangW0Q0021, HanWYS021}, we utilize the top-$k$ similarity search as validation method, adopting HR@10, HR@50, and R10@50 as evaluation metrics. The ground-truth results of top-$k$ similarity search are the exact top-$k$ similarity search results obtained when using traditional non-leaning based distance measures, including 
TP~\cite{shang2017trajectory}, DITA~\cite{shang2018dita}, LCRS~\cite{Yuan019}, and NetERP~\cite{KoideXI20}. Then, the basic idea when evaluating the effectiveness of similarity learning is to compare the top-$k$ results returned by the leaning-based methods with the top-$k$ results produced by the non-learning methods. Specifically, HR@$k$ denotes the top-$k$ hitting ratio that captures the degree of overlap between a top-$k$ result and the corresponding ground-truth result; and R$k$@$t$ is the top-$t$ recall for the top-$k$ ground truth that captures the fraction of the top-$k$ ground truth in the corresponding top-$t$ result. The closer HR@10, HR@50, and R10@50 are to 1, the higher the model effectiveness (i.e., similarity learning performance). 

\begin{table*}[tb]
\vspace{-5mm}
\caption{The Comparison of Similarity Learning on TP, DITA, LCRS, and NetERP Distances using Rome Dataset}
\vspace{-2.5mm}
\hspace{-5mm}
\begin{tabular}{p{1.4cm}<{\centering}|p{1.6cm}<{\centering}|p{0.82cm}<{\centering}p{0.82cm}<{\centering}p{0.86cm}<{\centering}|p{0.82cm}<{\centering}p{0.82cm}<{\centering}p{0.86cm}<{\centering}|p{0.82cm}<{\centering}p{0.82cm}<{\centering}p{0.86cm}<{\centering}|p{0.82cm}<{\centering}p{0.82cm}<{\centering}p{0.86cm}<{\centering}}
\hline
\multirow{2}{*}{Category} & \multirow{2}{*}{Methods} & \multicolumn{3}{c|}{TP [22]}            & \multicolumn{3}{c|}{DITA [26]}                & \multicolumn{3}{c|}{LCRS [44]}   & \multicolumn{3}{c}{NetERP [14]}  \\ \cline{3-14}
                          &                          & HR@10          & HR@50          & R10@50         & HR@10          & HR@50          & R10@50         & HR@10      & HR@50      & R10@50     & HR@10      & HR@50      & R10@50     \\ \hline
\multirow{4}{*}{\begin{tabular}[c]{@{}c@{}}Window\\ Guided \\ Baselines\end{tabular}}  
& NEUTRAJ$^\textit{w}$                    & 0.0976	&0.1499	&0.1775	&0.0898	&0.1417	&0.1756	&0.0405	&0.1552	&0.2488	&0.0053	&0.0397	&0.0723 \\
& Traj2SimVec$^\textit{w}$                    & 0.0552	&0.089	&0.0973	&0.0363	&0.0391	&0.0753	&0.0057	&0.026	&0.03138	&0.1191	&0.2235	&0.2728\\
& T3S$^\textit{w}$                    & 0.1098	&0.1863	&0.2228	&0.0893	&0.1426	&0.1823	&0.0669	&0.1766	&0.2910	&0.0123	&0.0512	&0.0871  \\
& GTS$^\textit{w}$                    & 0.1738	&0.3775	&0.4952	&0.0872	&0.1612	&0.2636	&0.1915	&0.2677	&0.4798	&0.0697	&0.1508	&0.1869 \\ \hline
\multirow{4}{*}{\begin{tabular}[c]{@{}c@{}}LSTM\\ Guided \\ Baselines\end{tabular}}  
& NEUTRAJ$^\textit{l}$                    &0.1225	&0.2177	&0.2613	&0.0932	&0.1499	&0.1950	&0.0864	&0.1981	&0.3308	&0.0172	&0.0608	&0.1004 \\
& Traj2SimVec$^\textit{l}$                    & 0.1108	&0.2287	&0.2712	&0.0544	&0.0772	&0.1336	&0.0992	&0.1350	&0.23309	&0.1151	&0.2205	&0.2787 \\
& T3S$^\textit{l}$                    &0.1195	&0.2092	&0.2508	&0.0931	&0.1494	&0.1931	&0.0805	&0.1930	&0.3209	&0.0156	&0.0582	&0.0969  \\
& GTS$^\textit{l}$                    & 0.1891	&0.4188	&0.5405	&0.0896	&0.1644	&0.2678	&0.2217	&0.2985	&0.5361	&0.0732	&0.1566	&0.2001  \\ \hline
\multirow{4}{*}{\begin{tabular}[c]{@{}c@{}}Our TMM\\ Guided \\ Baselines\end{tabular}}  
& NEUTRAJ$^\textit{t}$                   & 0.2092          & 0.4725         & 0.5986         & 0.0931         & 0.1692         & 0.2743         & 0.2606          & 0.3372          & 0.6088          & 0.0606          & 0.1254          & 0.2763  \\
& Traj2SimVec$^\textit{t}$                    & 0.2065          & 0.4654          & 0.5821          & 0.0891          & 0.1573          & 0.2477          & 0.2383          & 0.2899          & 0.5299          & 0.2067         & 0.2921          & 0.4711  \\
& T3S$^\textit{t}$                    & 0.2473         & 0.4994         & 0.5171          & 0.1876          & 0.2652          & 0.4729          & 0.2278          & 0.3098          & 0.4711          & 0.1217          & 0.2458          & 0.4608  \\
& GTS$^\textit{t}$                    & 0.3191          & 0.4229          & 0.6467          & 0.2148          & 0.3538          & 0.5226          & 0.2878          & 0.3185          & 0.5562          & 0.1935          & 0.2746          & 0.4177  \\ \hline
\multirow{1}{*}{Our Methods}  
& ST2Vec                    & \textbf{0.3834}      & \textbf{0.5051}	 &\textbf{0.7221}	&\textbf{0.2421}	 &\textbf{0.3689}	&\textbf{0.5614}	&\textbf{0.3178}	&\textbf{0.3942}	&\textbf{0.7244}	&\textbf{0.2117}	&\textbf{0.2967}	&\textbf{0.5117} \\ \hline
\end{tabular}
\label{tab:comparisonRome}
\end{table*}

\begin{table*}[tb]
\vspace{-3mm}
\caption{The Comparison of Similarity Learning on TP, DITA, LCRS, and NetERP Distances using Xi'an Dataset}
\vspace{-2.5mm}
\hspace{-5mm}
\begin{tabular}{p{1.4cm}<{\centering}|p{1.6cm}<{\centering}|p{0.82cm}<{\centering}p{0.82cm}<{\centering}p{0.86cm}<{\centering}|p{0.82cm}<{\centering}p{0.82cm}<{\centering}p{0.86cm}<{\centering}|p{0.82cm}<{\centering}p{0.82cm}<{\centering}p{0.86cm}<{\centering}|p{0.82cm}<{\centering}p{0.82cm}<{\centering}p{0.86cm}<{\centering}}
\hline
\multirow{2}{*}{Category} & \multirow{2}{*}{Methods} & \multicolumn{3}{c|}{TP [22]}            & \multicolumn{3}{c|}{DITA [26]}                & \multicolumn{3}{c|}{LCRS [44]}   & \multicolumn{3}{c}{NetERP [14]}  \\ \cline{3-14}
                          &                          & HR@10          & HR@50          & R10@50         & HR@10          & HR@50          & R10@50         & HR@10      & HR@50      & R10@50     & HR@10      & HR@50      & R10@50     \\ \hline
\multirow{4}{*}{\begin{tabular}[c]{@{}c@{}}Window\\ Guided \\ Baselines\end{tabular}}  
& NEUTRAJ$^\textit{w}$                    & 0.1353	&0.1946	&0.2369	&0.1326	&0.1953	&0.2397	&0.0742	&0.0996	&0.1240	&0.0401	&0.1856	&0.1878  \\
& Traj2SimVec$^\textit{w}$                    & 0.0689	&0.1154	&0.1446	&0.0247	&0.0628	&0.0749	&0.0103	&0.0166	&0.0157	&0.1148	&0.2185	&0.2299  \\
& T3S$^\textit{w}$                    & 0.1398	&0.2086	&0.2617	&0.1321	&0.1987	&0.2525	&0.0705	&0.0995	&0.1258	&0.0384	&0.1725	&0.1751  \\
& GTS$^\textit{w}$                    & 0.1640	&0.2679	&0.3748	&0.0763	&0.1470	&0.2444	&0.0253	&0.0569	&0.0928	&0.1496	&0.2277	&0.2801  \\ \hline
\multirow{4}{*}{\begin{tabular}[c]{@{}c@{}}LSTM\\ Guided \\ Baselines\end{tabular}}  
& NEUTRAJ$^\textit{l}$                    & 0.1763	&0.2352	&0.2879	&0.1331	&0.1854	&0.2384	&0.0746	&0.1324	&0.1637	&0.0573	&0.1861	&0.2106  \\
& Traj2SimVec$^\textit{l}$                   &0.1136	&0.1648	&0.2080	&0.0337	&0.0756	&0.1072	&0.0176	&0.0222	&0.0349	&0.1969	&0.3264	&0.3765  \\
& T3S$^\textit{l}$                    & 0.1908	&0.2582	&0.3222	&0.1355	&0.1899	&0.2529	&0.0734	&0.1409	&0.1755	&0.0603	&0.1803	&0.2099  \\
& GTS$^\textit{l}$                    &0.2995	&0.3941	&0.5125	&0.1254	&0.1727	&0.2818	&0.0607	&0.1912	&0.2474	&0.1627	&0.2465	&0.3456  \\ \hline
\multirow{4}{*}{\begin{tabular}[c]{@{}c@{}}Our TMM\\ Guided \\ Baselines\end{tabular}}  
& NEUTRAJ$^\textit{t}$                    &0.2169	&0.4892	&0.6222	&0.2197	&0.2612	&0.4228	&0.1143	&0.1928	&0.4869	&0.1820	&0.2742	&0.4368 \\
& Traj2SimVec$^\textit{t}$                &0.2310	&0.4288	&0.7799	&0.2035	&0.2329	&0.3735	&0.1158	&0.3783	&0.4708	&0.2232	&0.3336	&0.6268  \\
& T3S$^\textit{t}$                    &0.2545	&0.3870	&0.5341	&0.2319	&0.4049	&0.5397	&0.1286	&0.1663	&0.3197	&0.1630	&0.2943	&0.4461  \\
& GTS$^\textit{t}$                    &0.4190	&0.5363	&0.7937	&0.4086	&0.4011	&0.7778	&0.1049	&0.2375	&0.5235	&0.2318	&0.2939	&0.5112\\ \hline
\multirow{1}{*}{Our Methods}  
& ST2Vec                    & \textbf{0.4628} &\textbf{0.6014} &\textbf{0.8646} &\textbf{0.4128} &\textbf{0.5367}	&\textbf{0.8132}	&\textbf{0.1412}	&\textbf{0.2893}	&\textbf{0.6105}	&\textbf{0.3684}	&\textbf{0.4247}	&\textbf{0.7231} \\ \hline
\end{tabular}
\label{tab:comparisonXian}
\vspace{0mm}
\end{table*}

\noindent
\textbf{Competitors/Baselines.} We compare ST2Vec with all existing similarity learning methods, including NEUTRAJ~\cite{seed}, Traj2SimVec~\cite{subsimilar}, T3S~\cite{YangW0Q0021}, and GTS~\cite{HanWYS021}. Only the code for NEUTRAJ was available, while the code for the others are not. Note that, GTS has the state-of-the-art performance. Hence, we first carefully implemented Traj2SimVec, T3S, and GTS according to their descriptions. As our work is the first deep learning based method for spatio-temporal trajectory similarity leaning, for fairness of comparisons, we extend these competitors with time control, resulting in 12 baselines in three categories. The symbols $\textit{w}$, $\textit{l}$, and $\textit{t}$ are used to indicate the categories of baselines.
\begin{itemize}\setlength{\itemsep}{-\itemsep}
\item \textbf{Window-guided baselines ($*^\textit{w}$):} In this category, we distribute trajectories across discrete time slots and perform top-$k$ similarity queries in each slot, resulting in NEUTRAJ$^\textit{w}$, Traj2SimVec$^\textit{w}$, T3S$^\textit{w}$, and GTS$^\textit{w}$. 
\item \textbf{LSTM-guided baselines ($*^\textit{l}$):} In this category, we feed temporal trajectories directly into an LSTM model, resulting in NEUTRAJ$^\textit{l}$, Traj2SimVec$^\textit{l}$, T3S$^\textit{l}$, and GTS$^\textit{l}$. 
\item \textbf{Our TMM-guided baselines ($*^\textit{t}$):} In this category, we integrate our temporal trajectory embedding module (i.e., TMM) into the competitors, resulting in NEUTRAJ$^\textit{t}$, Traj2SimVec$^\textit{t}$, T3S$^\textit{t}$, and GTS$^\textit{t}$. 
\end{itemize}

\noindent
\textbf{Hyperparameters.} For ST2Vec, we use the UF strategy as the default; for all comparison methods, we use the SF strategy as the default. We split each data set into training, validation, and test sets in the ratio 3:1:6. The default value of $\lambda$ is set to 0.5. We set the spatial and temporal embedding dimensionalities to 128. The number of hidden LSTM units is 128. We set the batch size to 50. We tune their parameters to obtain the best performance. Moreover, we train the model using Adam~\cite{KingmaB14} with an initial learning rate of 0.001. Finally, we implemented ST2Vec in Python and Pytorch. All experiments were conducted on a server with an Intel Silver 4210R, 2.40GHz CPU, 64-GB RAM, and a GeForce GTX-2080 Ti 11G GPU. All implementation codes and corresponding datasets
have been released online$\footnote{\footnotesize Code and data available at https://github.com/ZJU-DBL/ST2Vec}$ for further studies.

\subsection{Model Effectiveness Study}
To demonstrate the model (i.e., similarity learning) effectiveness, we conduct top-$k$ similarity queries and compare the performance of ST2Vec with all 12 baseline approaches. Tables~\ref{tab:comparisonTdrive},~\ref{tab:comparisonRome}, and~\ref{tab:comparisonXian} list the results on the three datasets. From these results, we provide observations and analyses as follows.

We first observe that, our TMM-guided baselines significantly outperform the window-guided and LSTM-guided baselines, indicating that the proposed temporal trajectory embedding module is effective. This is because, although the window-based and LTSM-based methods might capture the temporal information to some extent, they ignore the continuous nature of time and periodic patterns, restricting their effectiveness. The second observation is that in the same category, GTS and ST2Vec outperform the other methods on all metrics. The main reason is that GTS and ST2Vec consider road network topology in spatial correlation modeling, while the other methods only capture the sequence features in free space and cannot embed the structural dependencies in road networks. The third observation is that ST2Vec achieves substantially better accuracy than GTS on all distance measures and all datasets. This reflects the fact that GTS targets POI-based trajectory similarity computation that disregards the actual travel paths between adjacent POIs. Given a target trajectory, the trajectories with the same neighbor POIs constitute its returned as its top-$k$ similarity querying results, although the movement paths of such trajectories can be very different from that of the target trajectory. In contrast, ST2Vec is designed for fine-grained trajectory similarity learning and considers both locations and travel paths between adjacent sample locations. Consequently, ST2Vec is capable of better similarity learning performance. 

\begin{table*}[]
\vspace{-5mm}
\caption{Model Scalability Evaluation with Varying Number of Trajectories to Perform Top-$k$ Similarity Computation}
\vspace{-2.5mm}
\hspace{-4mm}
\setlength{\tabcolsep}{1mm}{
\begin{tabular}{c|c|cccc|cccc|cccc|cccc}
\hline
\multirow{2}{*}{Datasets} & \multirow{2}{*}{Methods} & \multicolumn{4}{c|}{TP Distance}                                                                                                    & \multicolumn{4}{c|}{DITA Distance}                                                                                                  & \multicolumn{4}{c|}{LCRS Distance}                                                                                                  & \multicolumn{4}{c}{NetERP Distance}                                                                                                \\ \cline{3-18} 
                          &                          & \multicolumn{1}{c|}{10k}            & \multicolumn{1}{c|}{50k}             & \multicolumn{1}{c|}{100k}            & 200k            & \multicolumn{1}{c|}{10k}            & \multicolumn{1}{c|}{50k}             & \multicolumn{1}{c|}{100k}            & 200k            & \multicolumn{1}{c|}{10k}            & \multicolumn{1}{c|}{50k}             & \multicolumn{1}{c|}{100k}            & 200k            & \multicolumn{1}{c|}{10k}            & \multicolumn{1}{c|}{50k}             & \multicolumn{1}{c|}{100k}            & 200k            \\ \hline
\multirow{5}{*}{T-Drive}  & NEUTRAJ$^l$                  & \multicolumn{1}{c|}{27.81}          & \multicolumn{1}{c|}{131.39}          & \multicolumn{1}{c|}{261.16}          & 534.24          & \multicolumn{1}{c|}{23.75}          & \multicolumn{1}{c|}{135.18}          & \multicolumn{1}{c|}{258.04}          & 537.96          & \multicolumn{1}{c|}{31.10}          & \multicolumn{1}{c|}{127.03}          & \multicolumn{1}{c|}{261.04}          & 529.72          & \multicolumn{1}{c|}{25.13}          & \multicolumn{1}{c|}{127.22}          & \multicolumn{1}{c|}{257.48}          & 538.23          \\ \cline{2-18} 
                          & Traj2SimVec$^l$              & \multicolumn{1}{c|}{93.66}          & \multicolumn{1}{c|}{458.90}          & \multicolumn{1}{c|}{927.87}          & 1862.11         & \multicolumn{1}{c|}{92.10}          & \multicolumn{1}{c|}{454.28}          & \multicolumn{1}{c|}{926.25}          & 1865.86         & \multicolumn{1}{c|}{98.55}          & \multicolumn{1}{c|}{461.15}          & \multicolumn{1}{c|}{924.49}          & 1866.50         & \multicolumn{1}{c|}{93.34}          & \multicolumn{1}{c|}{456.74}          & \multicolumn{1}{c|}{928.39}          & 1858.86         \\ \cline{2-18} 
                          & T3S$^l$                      & \multicolumn{1}{c|}{30.52}          & \multicolumn{1}{c|}{146.94}          & \multicolumn{1}{c|}{275.91}          & 542.53          & \multicolumn{1}{c|}{33.80}          & \multicolumn{1}{c|}{142.58}          & \multicolumn{1}{c|}{276.96}          & 541.31          & \multicolumn{1}{c|}{34.99}          & \multicolumn{1}{c|}{147.42}          & \multicolumn{1}{c|}{279.50}          & 542.94          & \multicolumn{1}{c|}{29.77}          & \multicolumn{1}{c|}{148.35}          & \multicolumn{1}{c|}{279.40}          & 540.53          \\ \cline{2-18} 
                          & GTS$^l$                      & \multicolumn{1}{c|}{34.65}          & \multicolumn{1}{c|}{159.52}          & \multicolumn{1}{c|}{299.19}          & 602.52          & \multicolumn{1}{c|}{37.67}          & \multicolumn{1}{c|}{156.25}          & \multicolumn{1}{c|}{297.24}          & 597.64          & \multicolumn{1}{c|}{35.63}          & \multicolumn{1}{c|}{158.60}          & \multicolumn{1}{c|}{300.50}          & 607.09          & \multicolumn{1}{c|}{37.68}          & \multicolumn{1}{c|}{159.61}          & \multicolumn{1}{c|}{296.45}          & 606.36          \\ \cline{2-18} 
                          & ST2Vec                   & \multicolumn{1}{c|}{\textbf{30.32}} & \multicolumn{1}{c|}{\textbf{145.94}} & \multicolumn{1}{c|}{\textbf{293.35}} & \textbf{597.71} & \multicolumn{1}{c|}{\textbf{25.81}} & \multicolumn{1}{c|}{\textbf{146.65}} & \multicolumn{1}{c|}{\textbf{293.03}} & \textbf{596.56} & \multicolumn{1}{c|}{\textbf{29.38}} & \multicolumn{1}{c|}{\textbf{143.89}} & \multicolumn{1}{c|}{\textbf{297.92}} & \textbf{593.77} & \multicolumn{1}{c|}{\textbf{28.17}} & \multicolumn{1}{c|}{\textbf{147.26}} & \multicolumn{1}{c|}{\textbf{290.46}} & \textbf{598.88} \\ \hline
\multirow{5}{*}{Rome}     & NEUTRAJ$^l$                  & \multicolumn{1}{c|}{22.44}          & \multicolumn{1}{c|}{97.01}           & \multicolumn{1}{c|}{191.55}          & 388.27          & \multicolumn{1}{c|}{21.39}          & \multicolumn{1}{c|}{101.70}          & \multicolumn{1}{c|}{192.50}          & 387.27          & \multicolumn{1}{c|}{25.46}          & \multicolumn{1}{c|}{94.97}           & \multicolumn{1}{c|}{196.04}          & 390.83          & \multicolumn{1}{c|}{22.54}          & \multicolumn{1}{c|}{98.15}           & \multicolumn{1}{c|}{192.83}          & 386.96          \\ \cline{2-18} 
                          & Traj2SimVec$^l$              & \multicolumn{1}{c|}{81.22}          & \multicolumn{1}{c|}{421.22}          & \multicolumn{1}{c|}{877.58}          & 1801.12         & \multicolumn{1}{c|}{81.52}          & \multicolumn{1}{c|}{425.17}          & \multicolumn{1}{c|}{882.12}          & 1800.89         & \multicolumn{1}{c|}{77.31}          & \multicolumn{1}{c|}{422.88}          & \multicolumn{1}{c|}{882.22}          & 1802.69         & \multicolumn{1}{c|}{76.34}          & \multicolumn{1}{c|}{418.81}          & \multicolumn{1}{c|}{873.00}          & 1796.19         \\ \cline{2-18} 
                          & T3S$^l$                      & \multicolumn{1}{c|}{24.54}          & \multicolumn{1}{c|}{100.31}          & \multicolumn{1}{c|}{199.38}          & 395.73          & \multicolumn{1}{c|}{23.51}          & \multicolumn{1}{c|}{97.26}           & \multicolumn{1}{c|}{200.27}          & 394.36          & \multicolumn{1}{c|}{27.74}          & \multicolumn{1}{c|}{96.67}           & \multicolumn{1}{c|}{195.80}          & 392.66          & \multicolumn{1}{c|}{20.94}          & \multicolumn{1}{c|}{97.26}           & \multicolumn{1}{c|}{195.04}          & 394.57          \\ \cline{2-18} 
                          & GTS$^l$                      & \multicolumn{1}{c|}{23.54}          & \multicolumn{1}{c|}{104.27}          & \multicolumn{1}{c|}{196.52}          & 395.62          & \multicolumn{1}{c|}{24.19}          & \multicolumn{1}{c|}{106.97}          & \multicolumn{1}{c|}{198.63}          & 394.01          & \multicolumn{1}{c|}{27.51}          & \multicolumn{1}{c|}{108.87}          & \multicolumn{1}{c|}{199.64}          & 398.01          & \multicolumn{1}{c|}{27.21}          & \multicolumn{1}{c|}{101.19}          & \multicolumn{1}{c|}{193.42}          & 393.58          \\ \cline{2-18} 
                          & ST2Vec                   & \multicolumn{1}{c|}{\textbf{21.66}} & \multicolumn{1}{c|}{\textbf{99.34}}  & \multicolumn{1}{c|}{\textbf{194.10}} & \textbf{392.46} & \multicolumn{1}{c|}{\textbf{23.64}} & \multicolumn{1}{c|}{\textbf{102.53}} & \multicolumn{1}{c|}{\textbf{198.55}} & \textbf{393.24} & \multicolumn{1}{c|}{\textbf{16.89}} & \multicolumn{1}{c|}{\textbf{95.83}}  & \multicolumn{1}{c|}{\textbf{198.12}} & \textbf{394.65} & \multicolumn{1}{c|}{\textbf{16.73}} & \multicolumn{1}{c|}{\textbf{99.37}}  & \multicolumn{1}{c|}{\textbf{198.69}} & \textbf{392.71} \\ \hline
\end{tabular}}
\label{tab:scalability}
\end{table*}

\begin{figure*} [tb]
	\centering
	\includegraphics[width=0.6\textwidth]{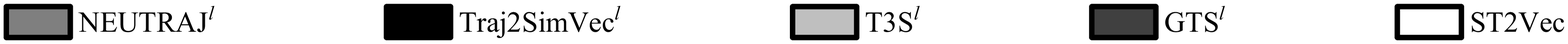}\\
	\vspace{-1.5mm}
		\subfigure[T-drive/TP]{
	    \includegraphics[width=4.3cm,height=2.5cm]{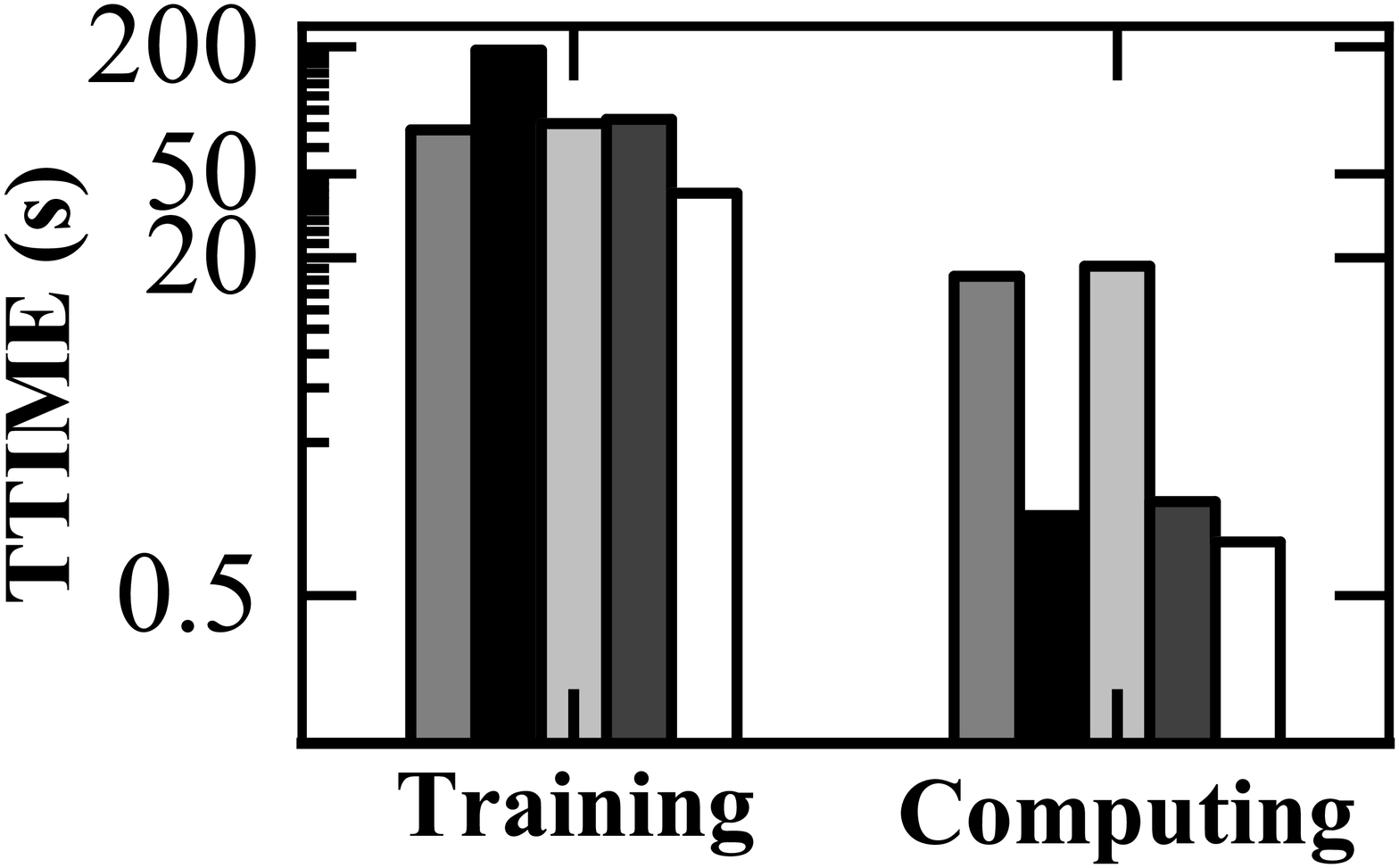}}
	\subfigure[T-drive/DITA]{
		\includegraphics[width=4.3cm,height=2.5cm]{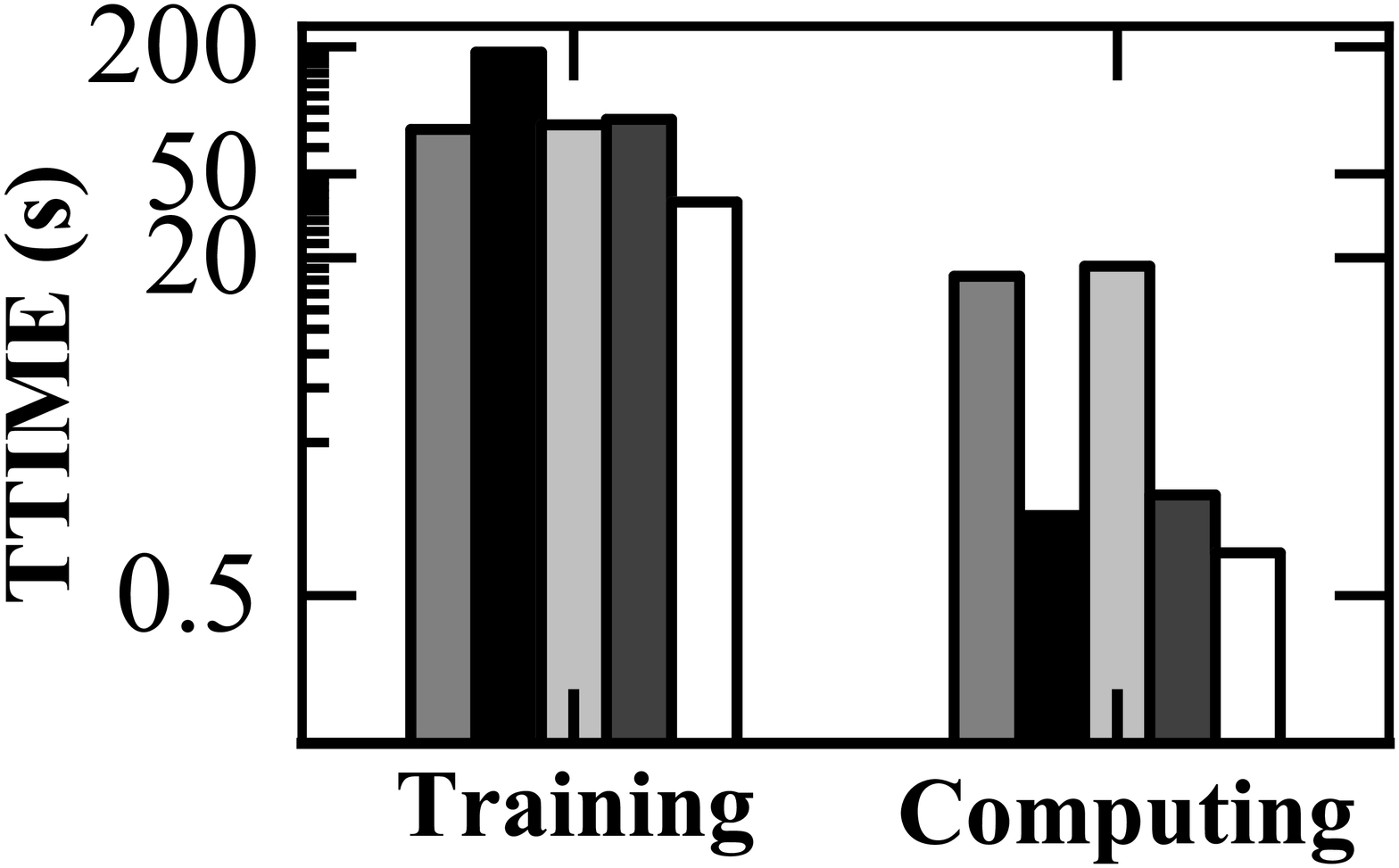}}
	\subfigure[T-drive/LCRS]{
		\includegraphics[width=4.3cm,height=2.5cm]{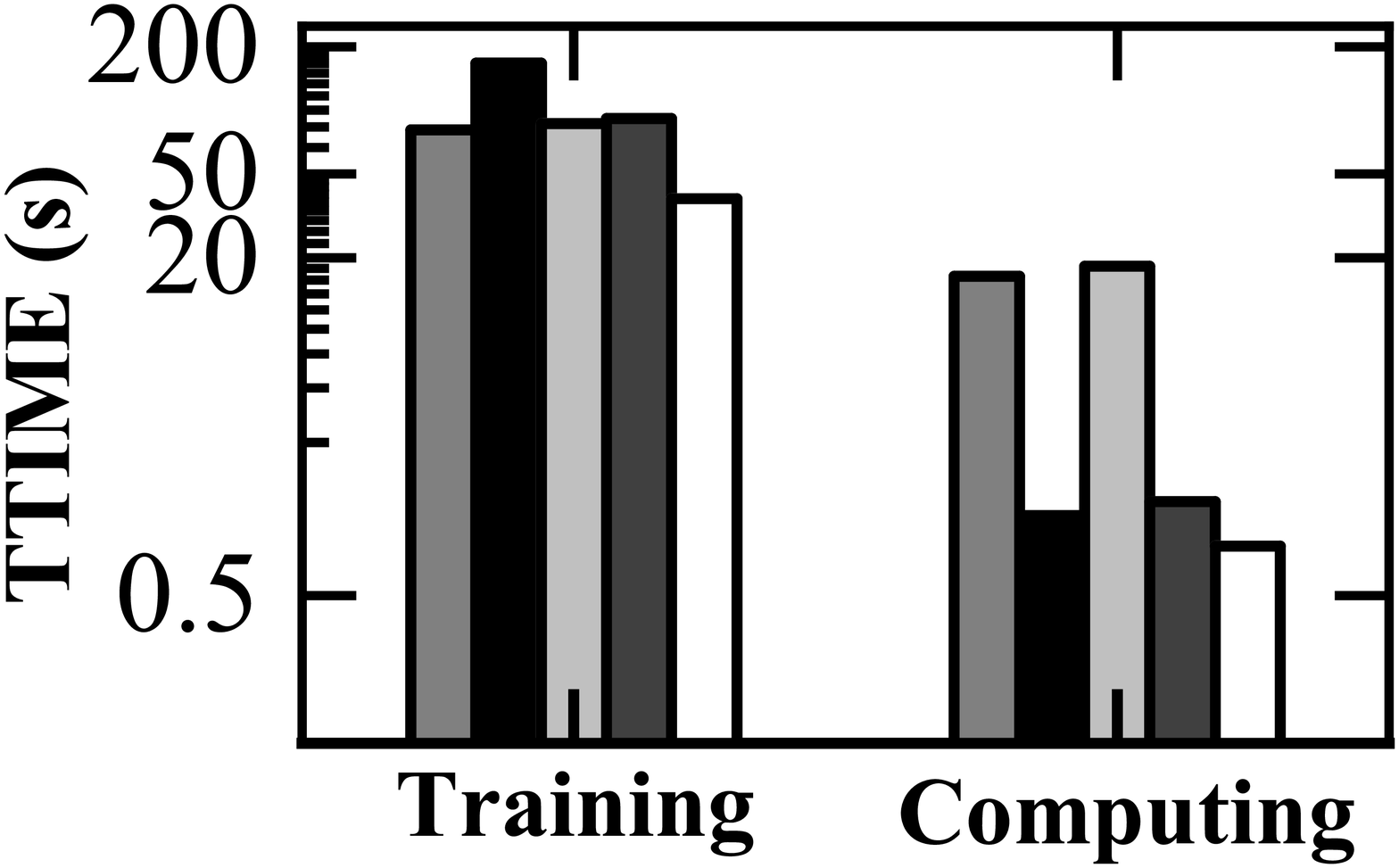}}
	\subfigure[T-drive/NetERP]{
		\includegraphics[width=4.3cm,height=2.5cm]{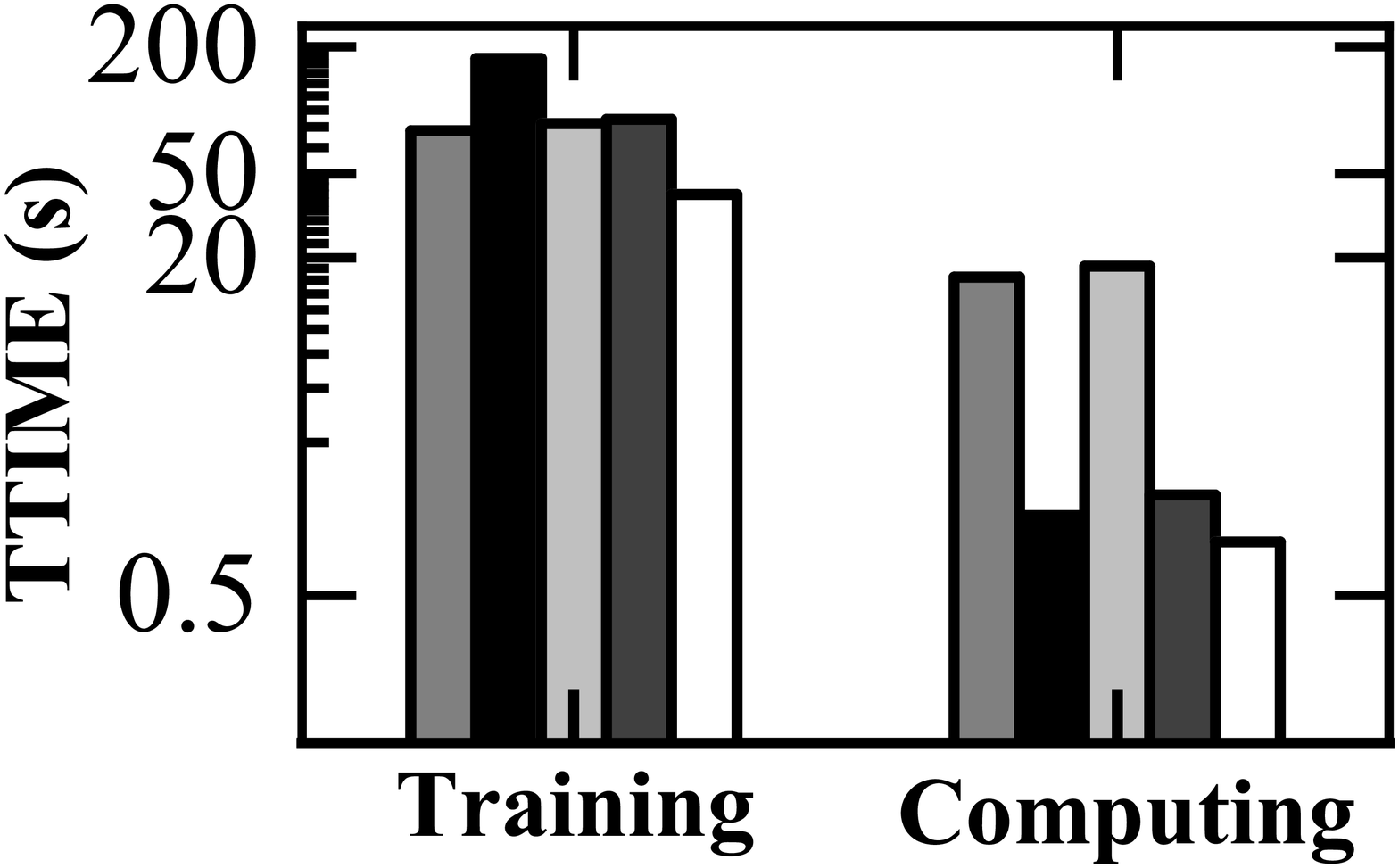}} \\
	\centering
	\vspace{-2mm}
		\subfigure[Rome/TP]{
	    \includegraphics[width=4.3cm,height=2.5cm]{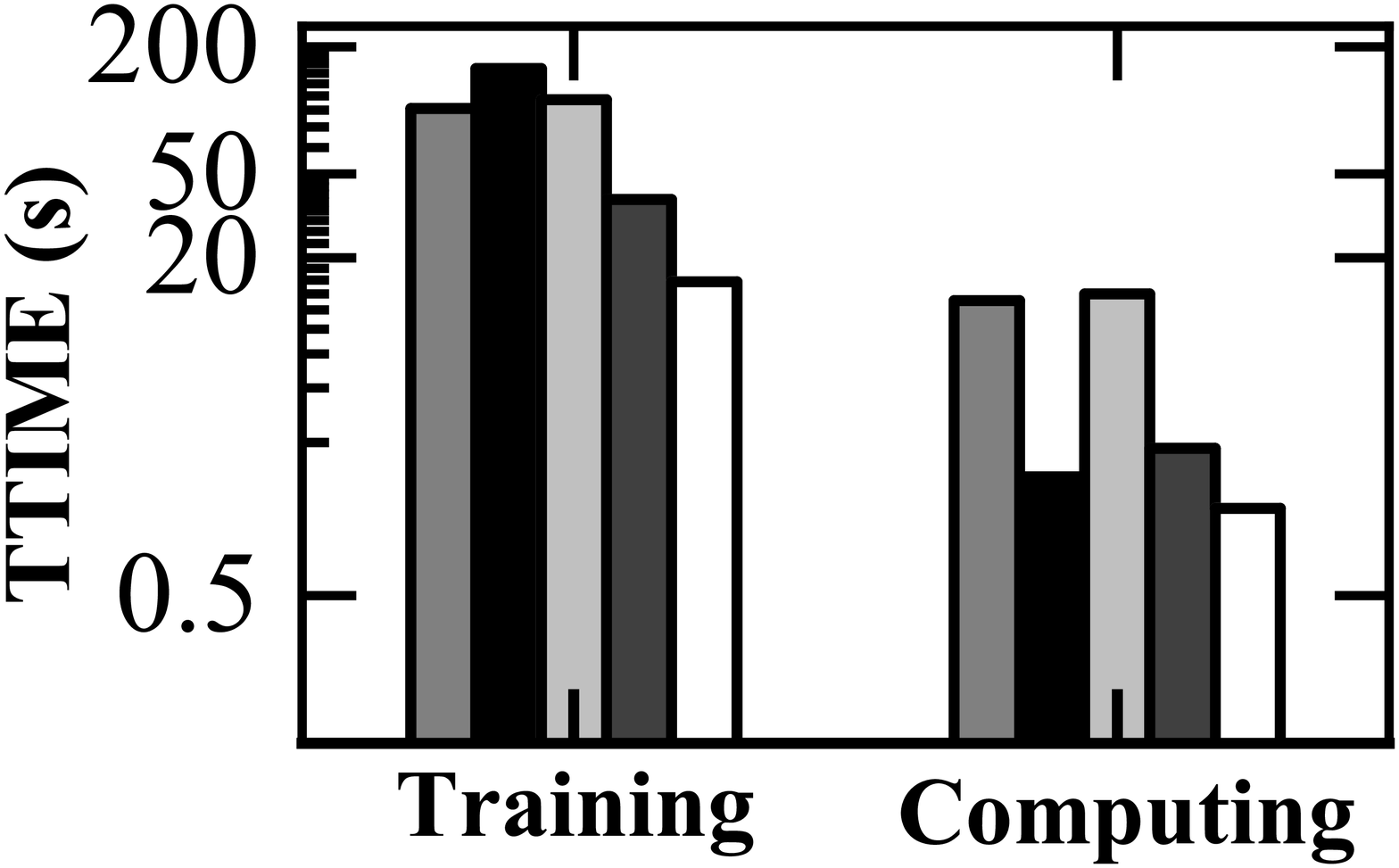}}
	\subfigure[Rome/DITA]{
		\includegraphics[width=4.3cm,height=2.5cm]{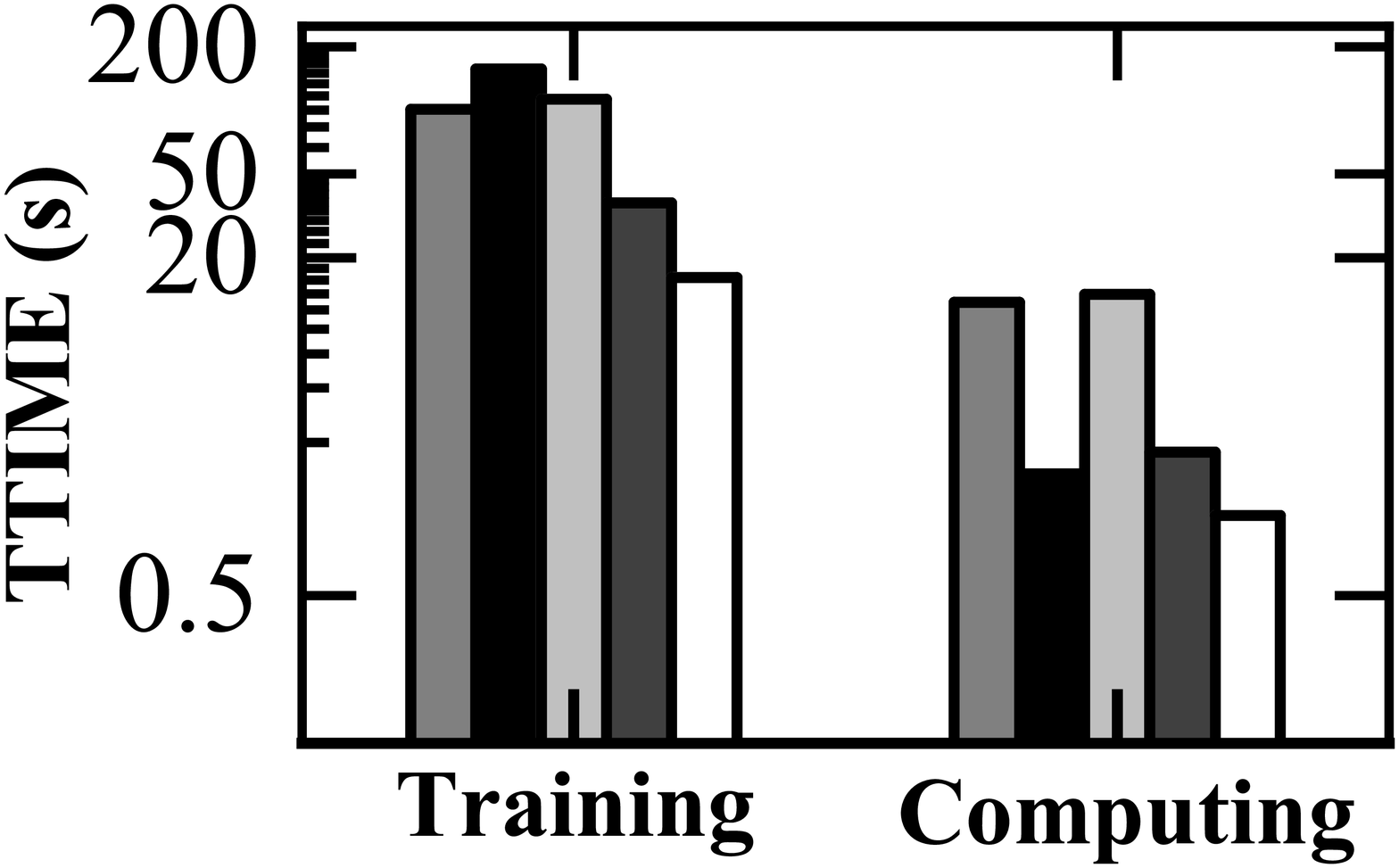}}
	\subfigure[Rome/LCRS]{
		\includegraphics[width=4.3cm,height=2.5cm]{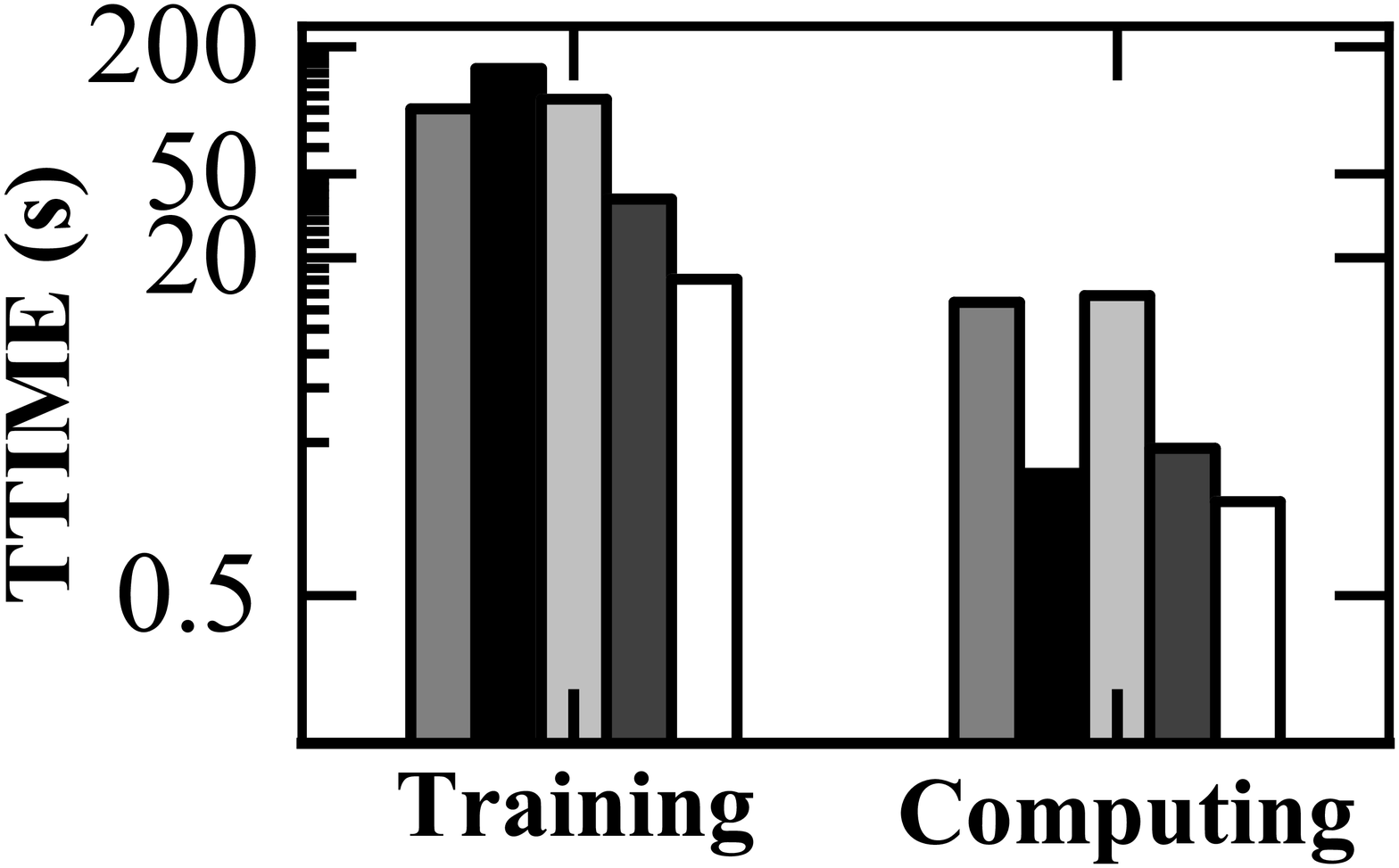}}
	\subfigure[Rome/NetERP]{
		\includegraphics[width=4.3cm,height=2.5cm]{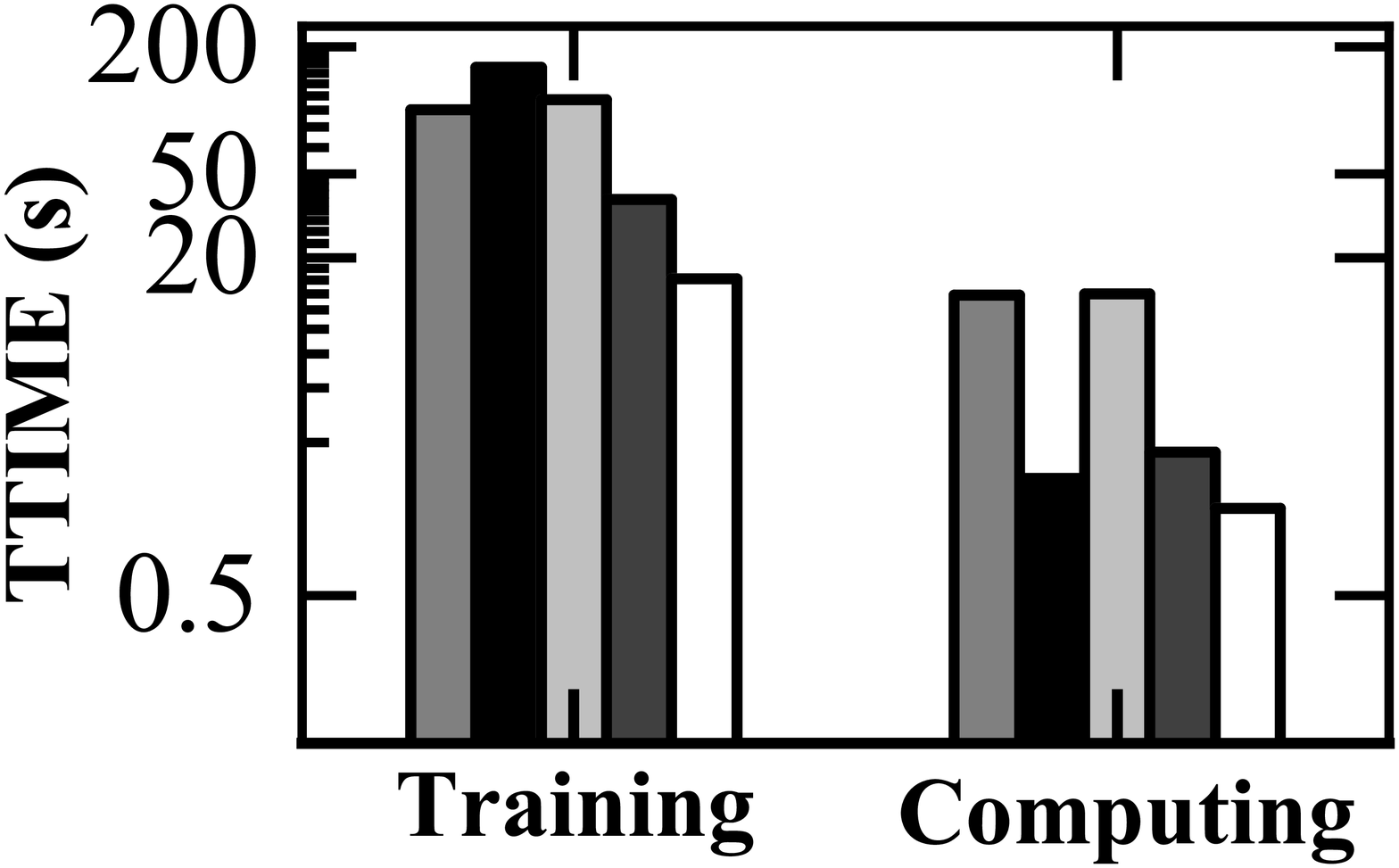}} \\
	\vspace{-2mm}
	\caption{Model Efficiency Evaluation on Offline Model Training and Online Computing Phases}
	\label{fig:efficiency} 
	\vspace{-2mm}
\end{figure*}

\subsection{Model Efficiency Study}
Next, we study the model efficiency in terms of both offline model training (denoted as training, with the unit seconds/epoch) and online computing (denoted as computing, with the unit seconds/4k trajectories). Fig.~\ref{fig:efficiency} shows the results on T-Drive and Rome. Note that the scale of the y-axis is logarithmic due to the significant performance differences. The results on Xi'an are similar and are omitted for brevity. We only compare ST2Vec with the LSTM-guided baselines because they outperform the window-guided baselines and because the TMM-guided baselines are essentially based on our TMM module. 

As can be seen, ST2Vec has good performance for both training and computing. Consider the results for T-drive as an example. During the training phase, ST2Vec finishes each epoch within 40 seconds and runs two times faster than NEUTRAJ, T3S, and GTS, and five times faster than Traj2SimVec. In terms of similarity computation (i.e., testing), we measure the total running time of each method on the test data. Here, ST2Vec also exhibits superior performance (i.e., within 1 second) and is 20 times faster than NEUTRAJ and T3S and two times faster than Traj2Sim and GTS. 

\subsection{Model Scalability Study}
Next, we explore model scalability when varying the number of trajectories from 10k to 200k. Table~\ref{tab:scalability} shows the results when learning four distance measures on T-Drive and Rome. The results on Xi'an are omitted because they yield similar observations. 

As can be observed, ST2Vec offers the best scalability for learning-based trajectory similarity computation due to three observations. First, the running time increases with the cardinality. Second, ST2Vec offers substantial performance improvements over the existing methods. Third, the performance of ST2Vec is affected less by an increase in cardinality than are the four baselines. Consequently, ST2Vec is capable of large-scale trajectory similarity computation.

\subsection{Parameter Sensitivity Study}
Further, we evaluate the sensitivity of ST2Vec to assess its robustness. Specifically, we consider the effects on the model performance of the training data size, the number of triplets $N$ constructed for each trajectory, and the spatio-temporal weight $\lambda$. We report results for T-drive only; Rome and Xi'an  yield similar observations. 

\begin{figure*} [tb]
	\centering
	\hspace{-0.25cm}
		\subfigure[T-drive/TP]{
	    \includegraphics[width=4.3cm,height=3cm]{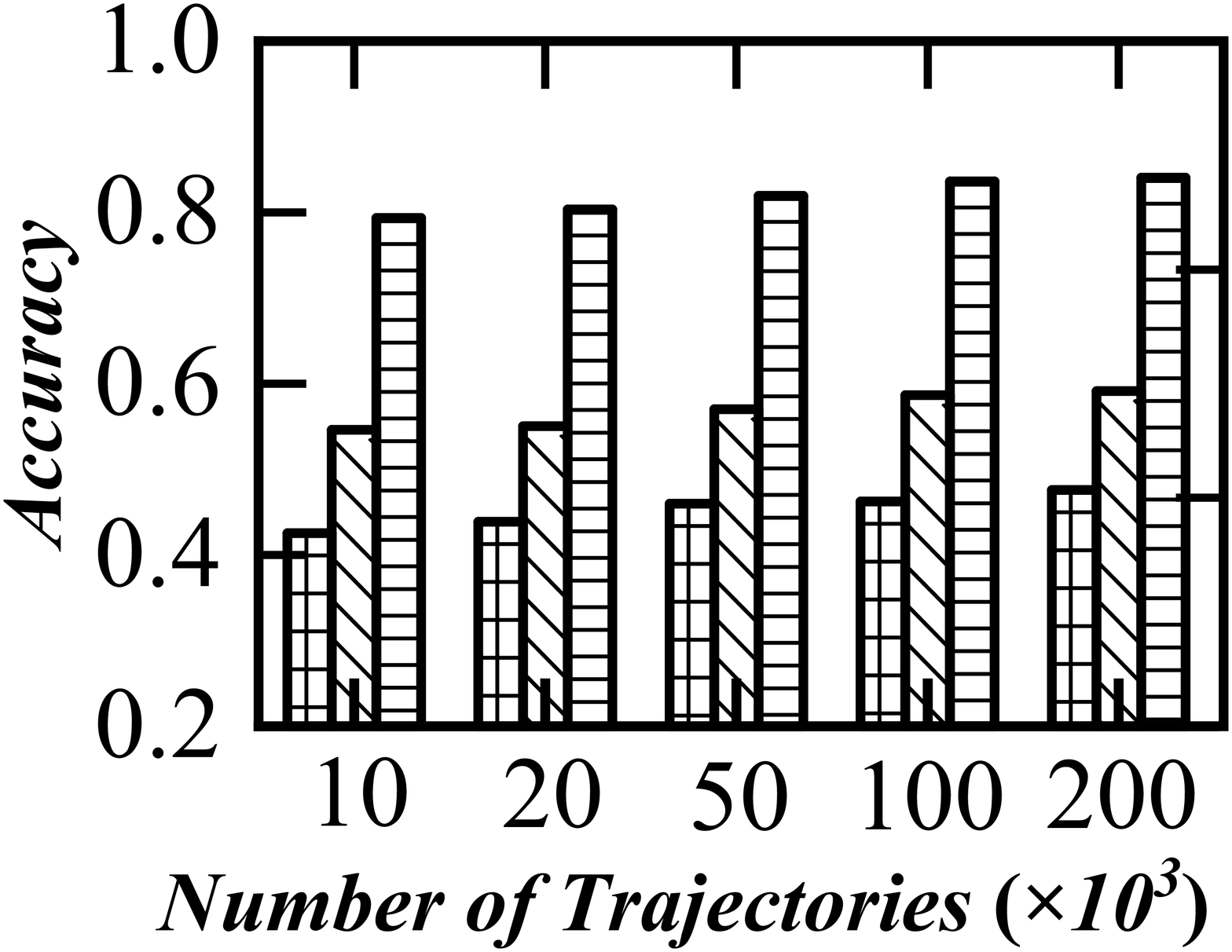}}
	\subfigure[T-drive/DITA]{
		\includegraphics[width=4.3cm,height=3cm]{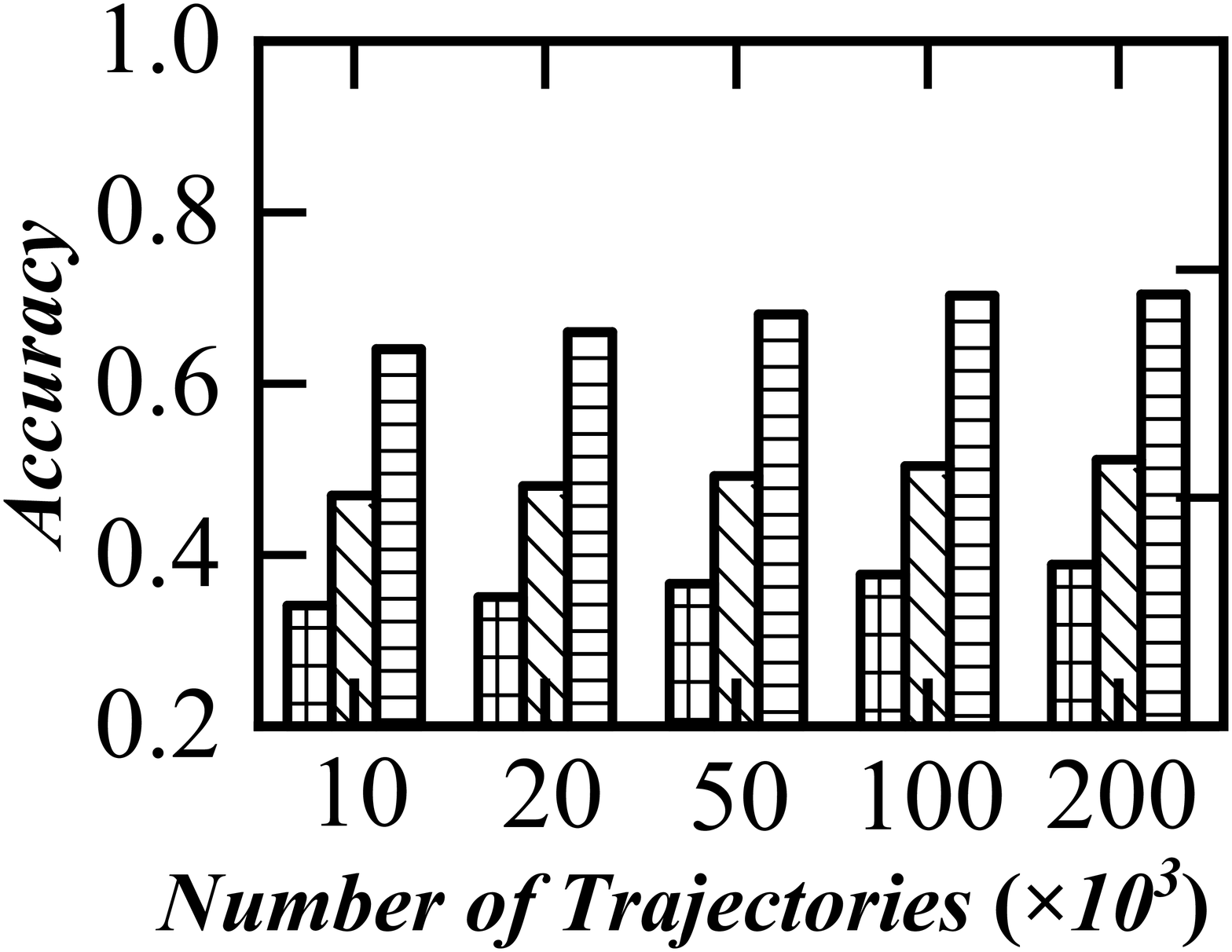}}
	\subfigure[T-drive/LCRS]{
		\includegraphics[width=4.3cm,height=3cm]{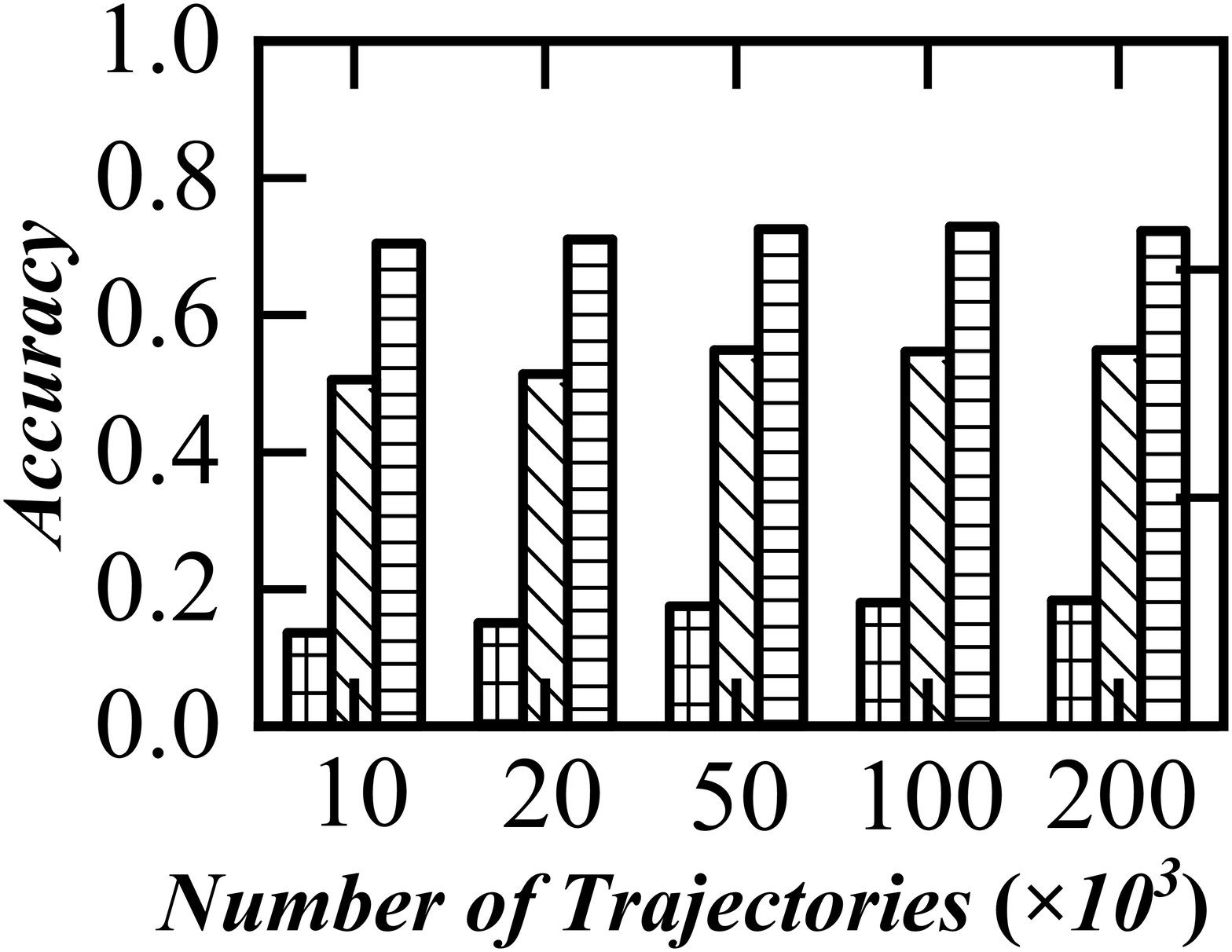}}
	\subfigure[T-drive/NetERP]{
		\includegraphics[width=4.3cm,height=3cm]{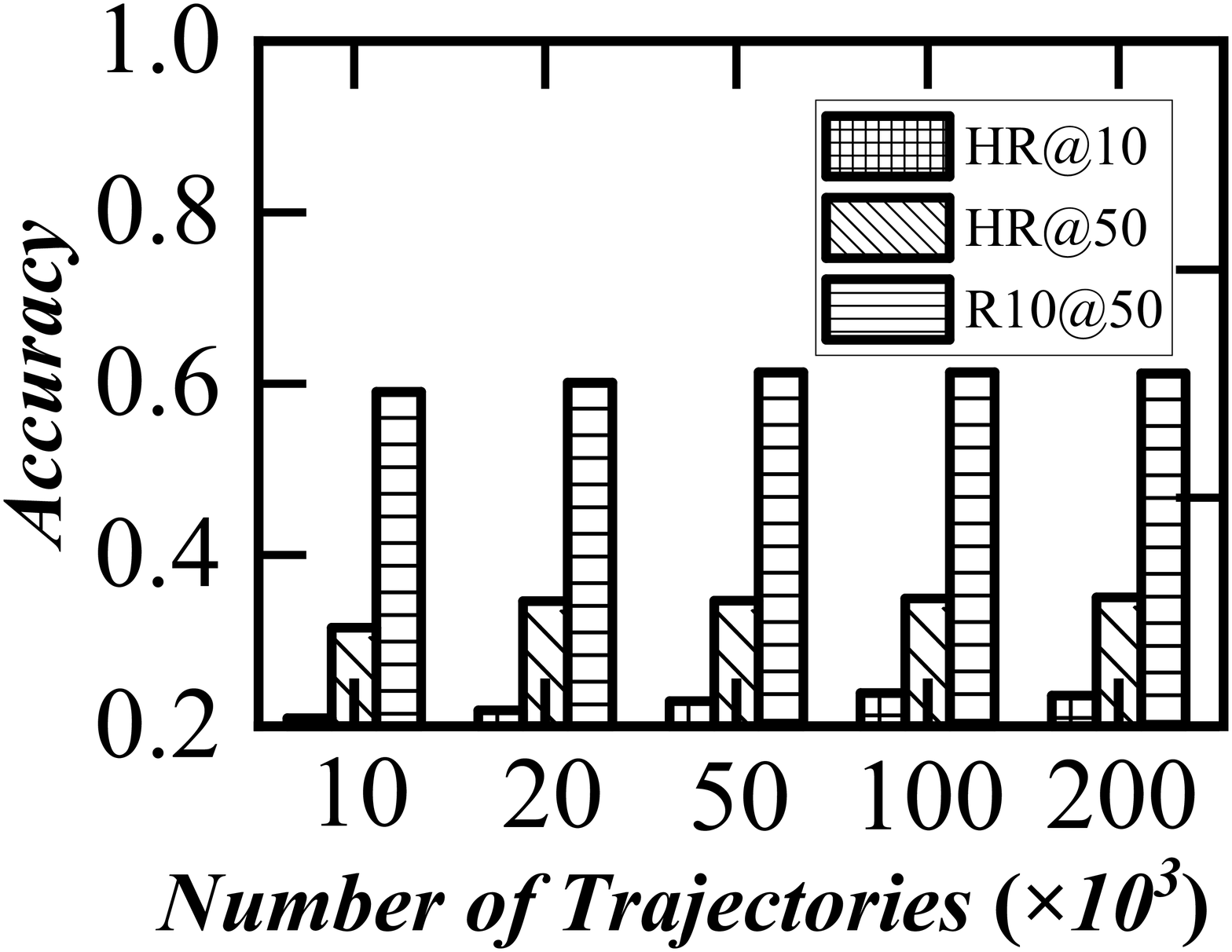}} \\
	\vspace{-2mm}
	\caption{Performance of ST2Vec under Varying Training Data Size}
	\label{fig:ModelAnalysisofsize}
	\vspace{-3mm}
\end{figure*}

\noindent
\textbf{Sensitivity to $datasize$.} First, we investigate the effect of the number of training trajectories on the performance of ST2Vec. Fig.~\ref{fig:ModelAnalysisofsize} shows the similarity learning performance (i.e., HR@10, HR@50, R10@50) for the four measures when varying the training data size from 10k to 200k. As can be observed, ST2Vec exhibits stable performance. 

\begin{figure*} [tb]
	\centering
	\hspace{-0.25cm}
		\subfigure[T-drive/TP]{
	    \includegraphics[width=4.3cm,height=3cm]{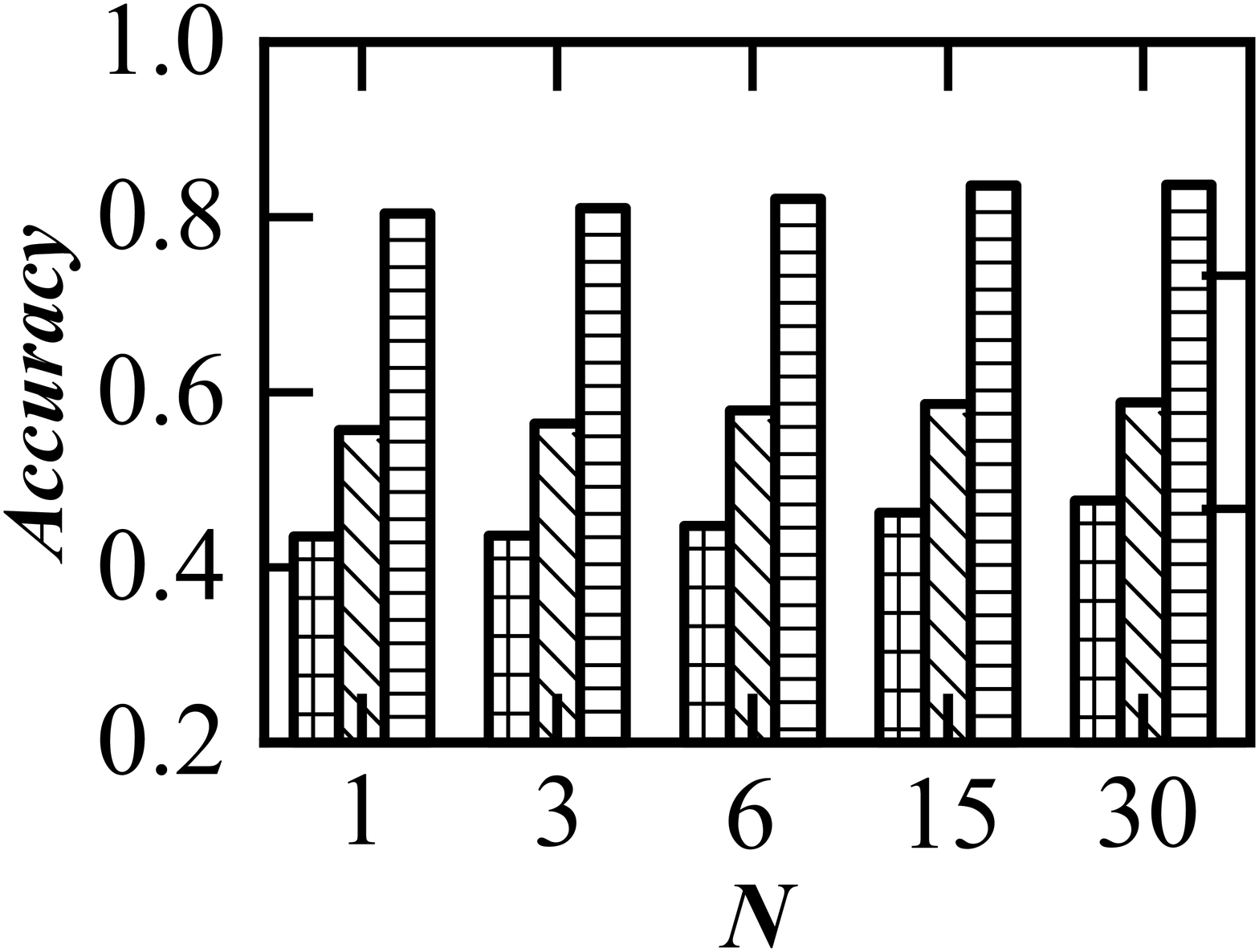}}
	\subfigure[T-drive/DITA]{
		\includegraphics[width=4.3cm,height=3cm]{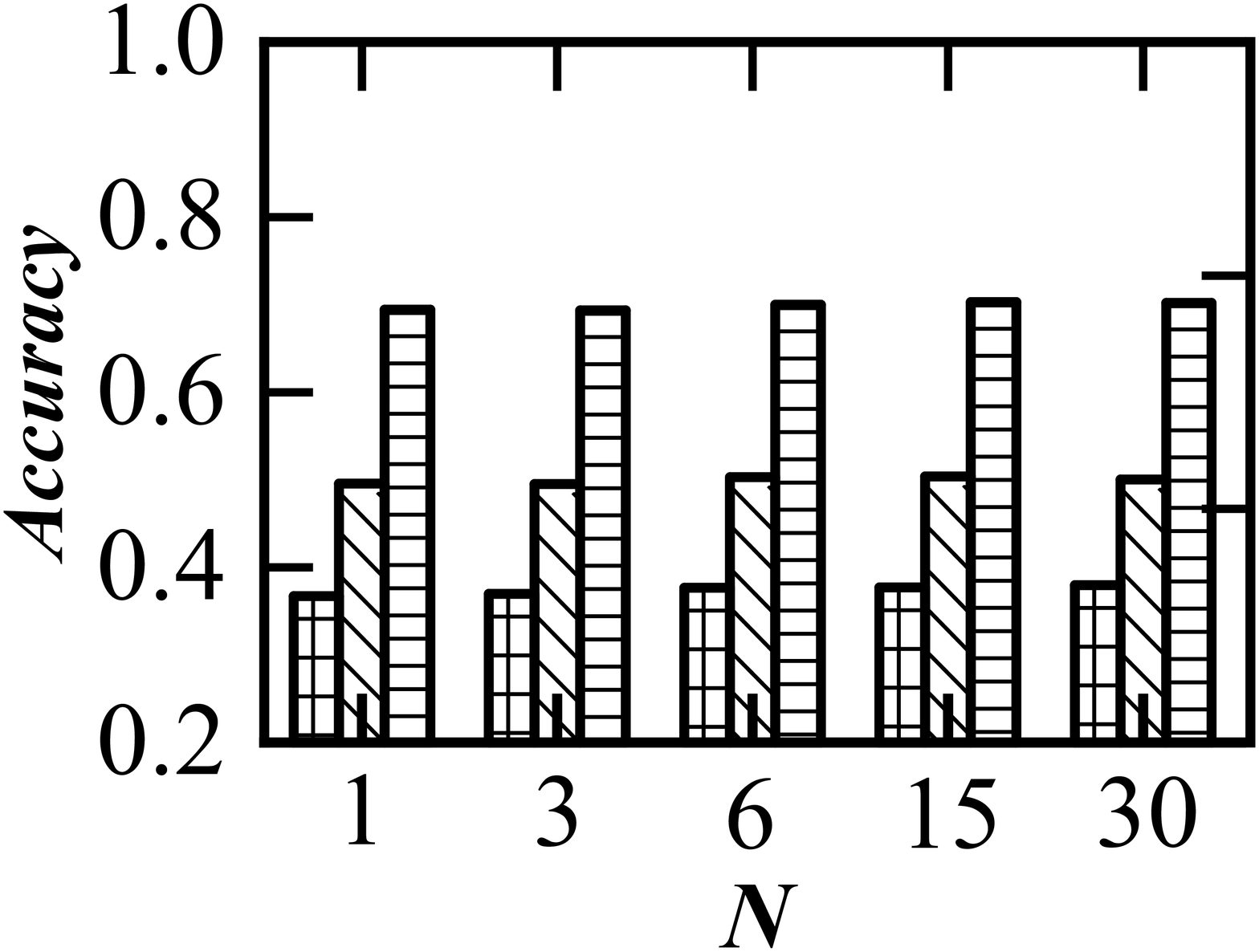}}
	\subfigure[T-drive/LCRS]{
		\includegraphics[width=4.3cm,height=3cm]{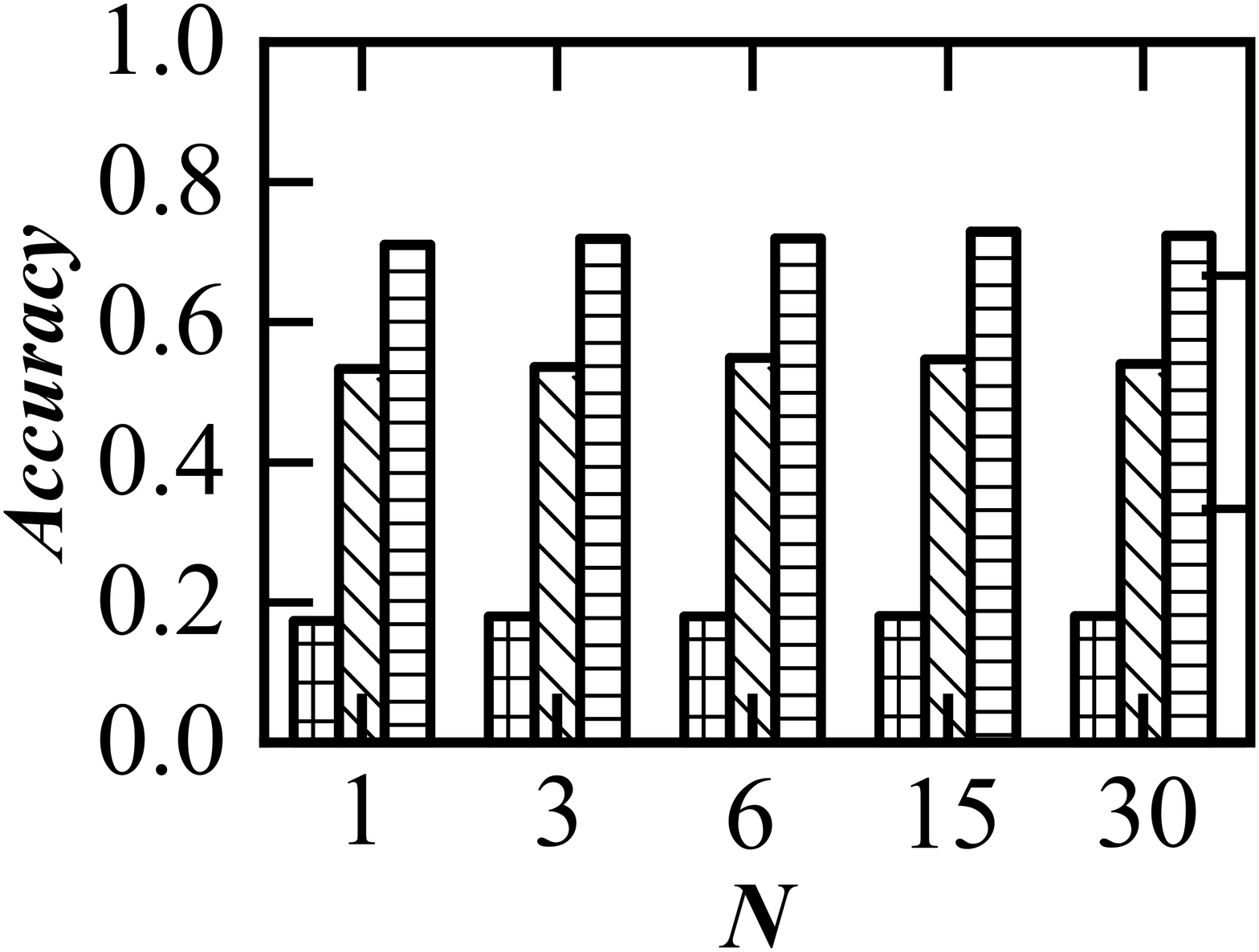}}
	\subfigure[T-drive/NetERP]{
		\includegraphics[width=4.3cm,height=3cm]{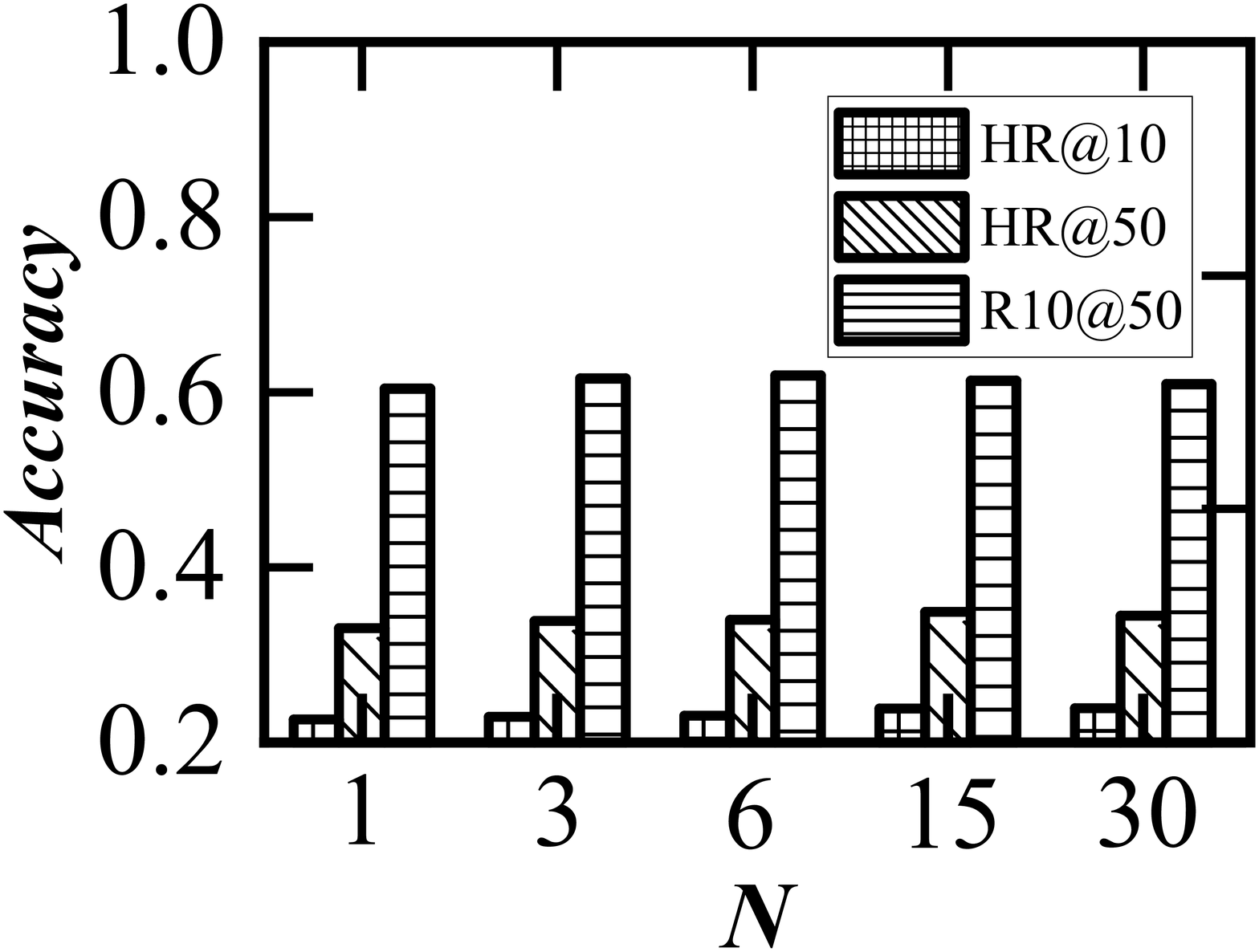}} \\
	\vspace{-2mm}
	\caption{Performance of ST2Vec under Varying Number of Triplets $N$ for Each Trajectory}
	\label{fig:ModelAnalysisofk}
	\vspace{-3mm} 
\end{figure*}

\noindent
\textbf{Sensitivity to $N$.} Second, we investigate model robustness when varying $k$ for constructing training samples. Here, we randomly sample 10k trajectories from T-Drive. Then, for each trajectory, we get its 1, 3, 6, 15, and 30 most similar/dissimilar trajectories to construct similarity triplets. Fig.~\ref{fig:ModelAnalysisofk} plots the results using four distance measures. As can be observed, HR@10, HR@50, and R10@50 all increase slightly, which offers evidence that ST2Vec is capable of achieving good performance even with limited training samples.

\begin{figure*} [tb]
	\centering
	\hspace{-0.25cm}
		\subfigure[T-drive/TP]{
	    \includegraphics[width=4.3cm,height=3cm]{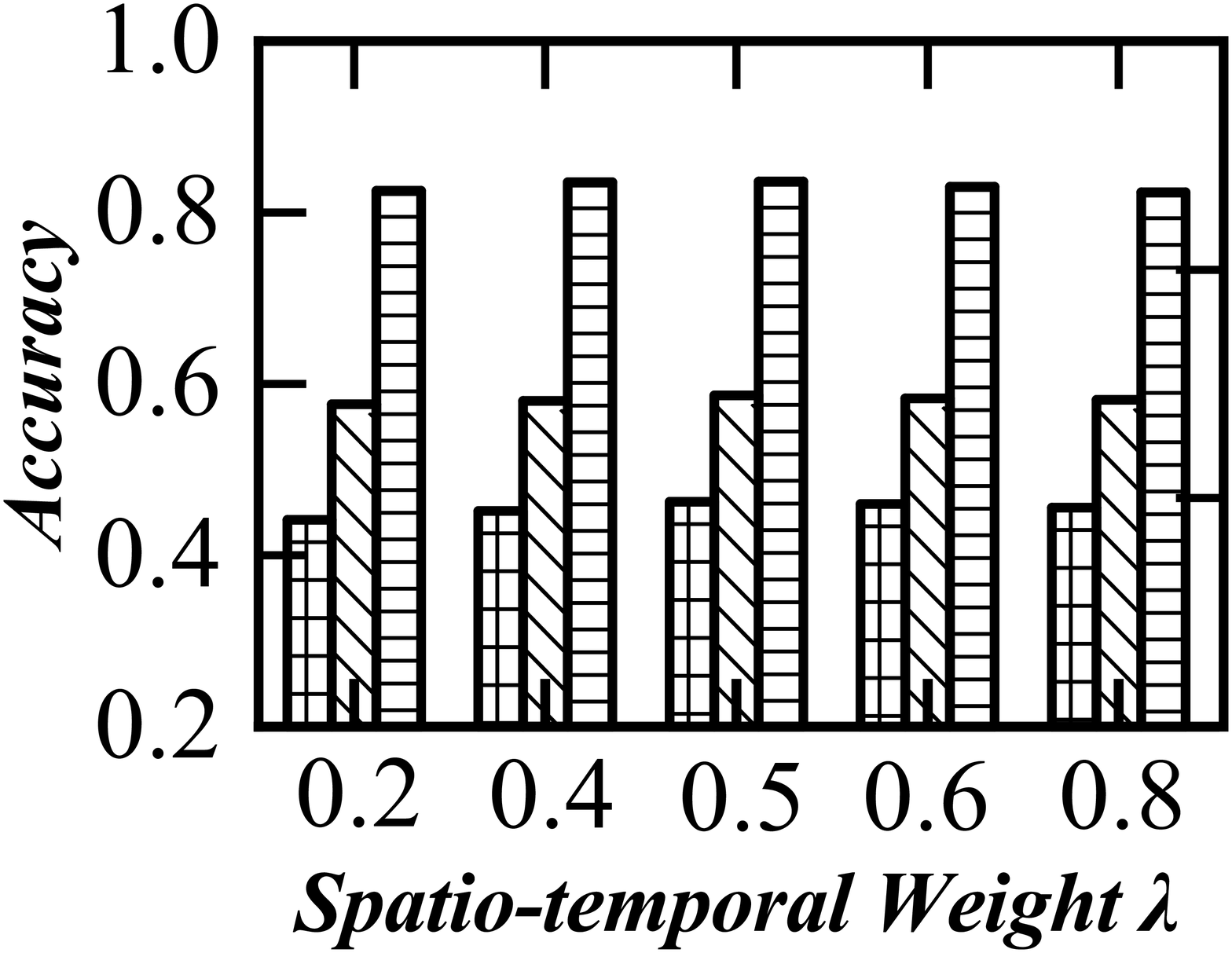}}
	\subfigure[T-drive/DITA]{
		\includegraphics[width=4.3cm,height=3cm]{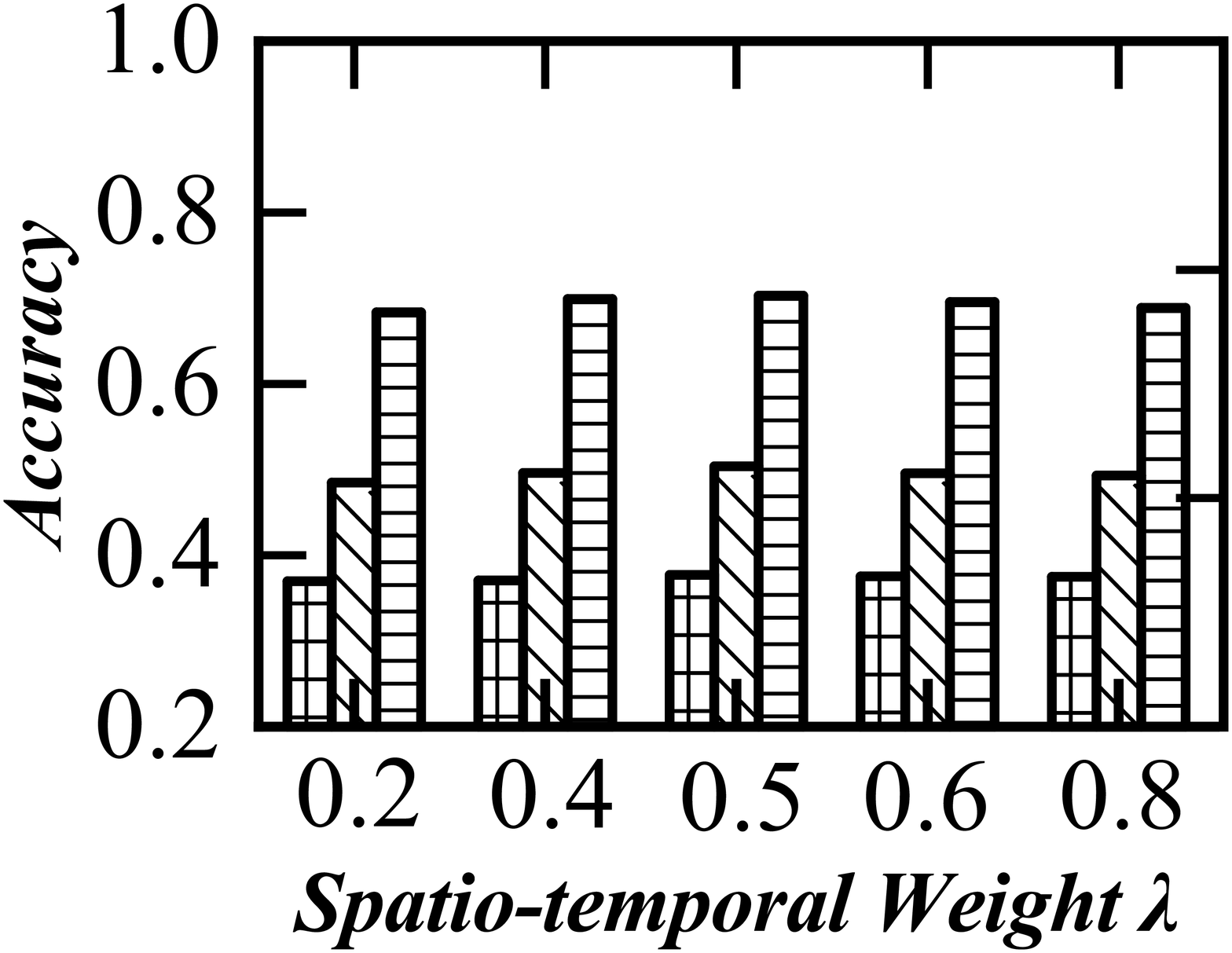}}
	\subfigure[T-drive/LCRS]{
		\includegraphics[width=4.3cm,height=3cm]{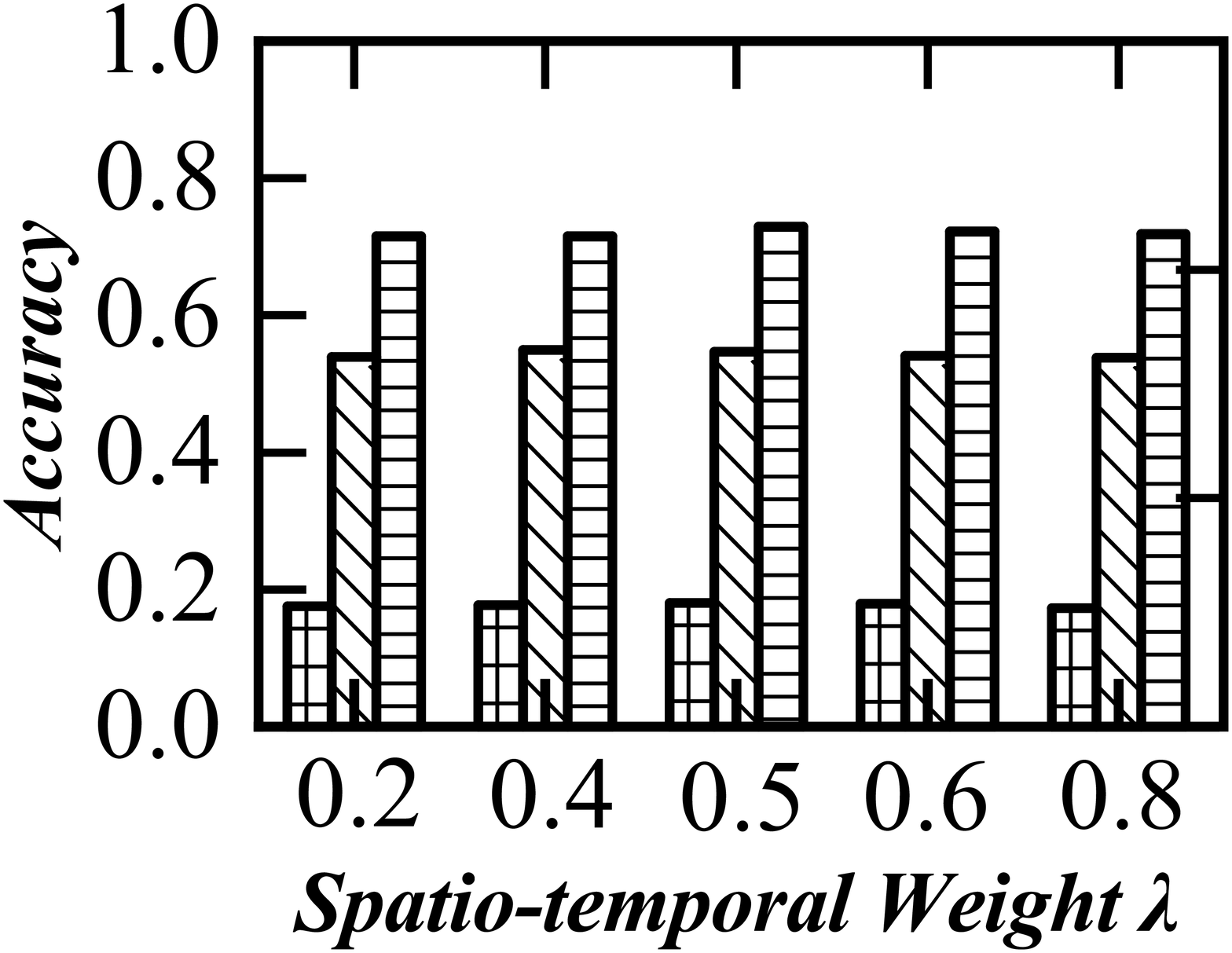}}
	\subfigure[T-drive/NetERP]{
		\includegraphics[width=4.3cm,height=3cm]{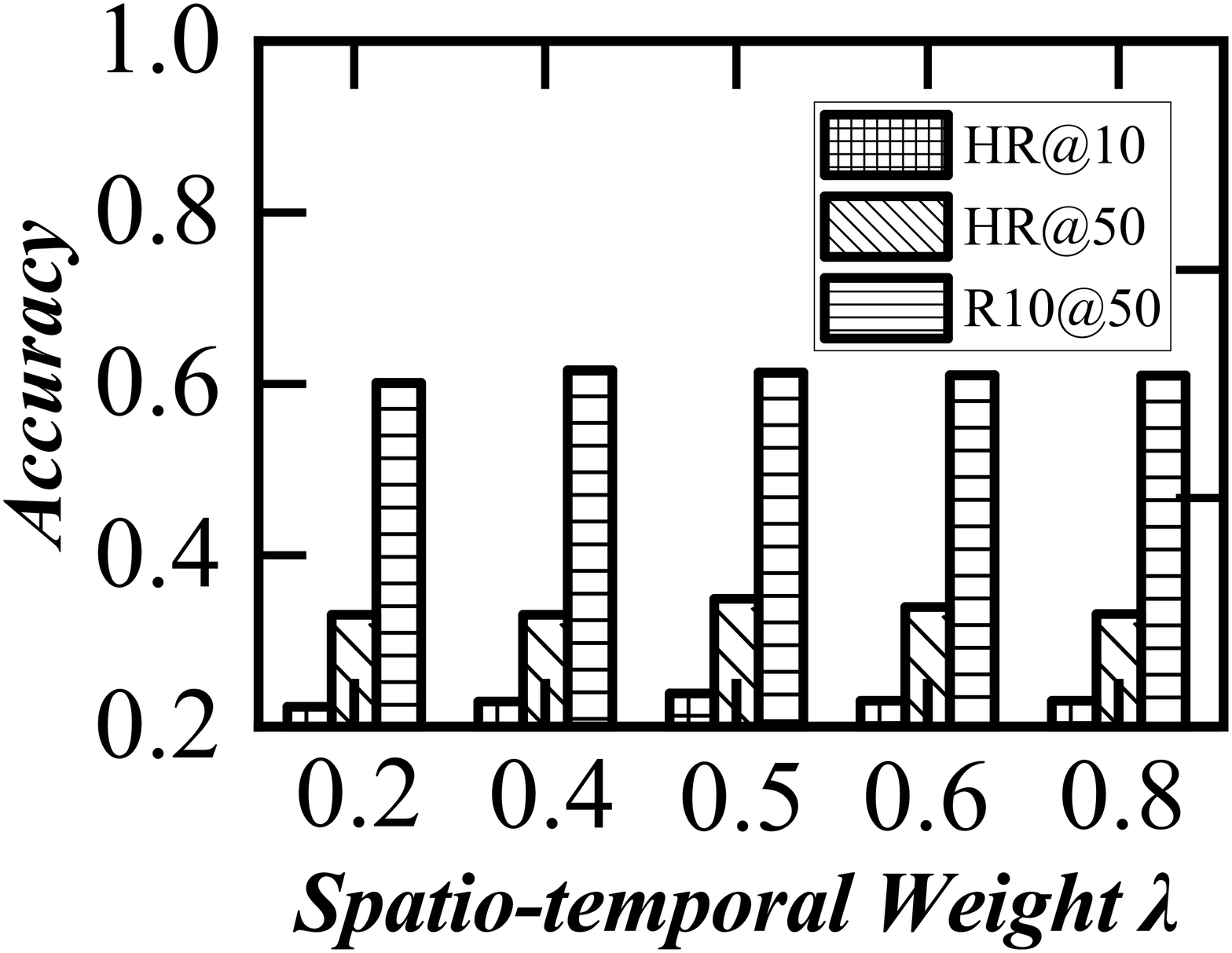}} \\
	\vspace{-2mm}
	\caption{Performance of ST2Vec under Varying Spatio-Temporal Weight $\lambda$}
	\label{fig:ModelAnalysisoflambda}
	\vspace{-3mm}
\end{figure*}

\noindent
\textbf{Sensitivity to $\lambda$.} Finally, we perform a  sensitivity analysis of the spatio-temporal weight $\lambda$ used in Eq. 1. When $\lambda$ = 1, only the spatial domain is considered, and when $\lambda$ = 0, the similar computation considers the temporal domain only. Fig.~\ref{fig:ModelAnalysisoflambda} shows that HR@10, HR@50, and R10@50 performance are stable across different settings of $\lambda$, indicating that ST2Vec works well with different $\lambda$ preferences. 

\begin{figure}[tb]
	\centering
	\subfigure[T-Drive]{
		\includegraphics[width=0.23\textwidth]{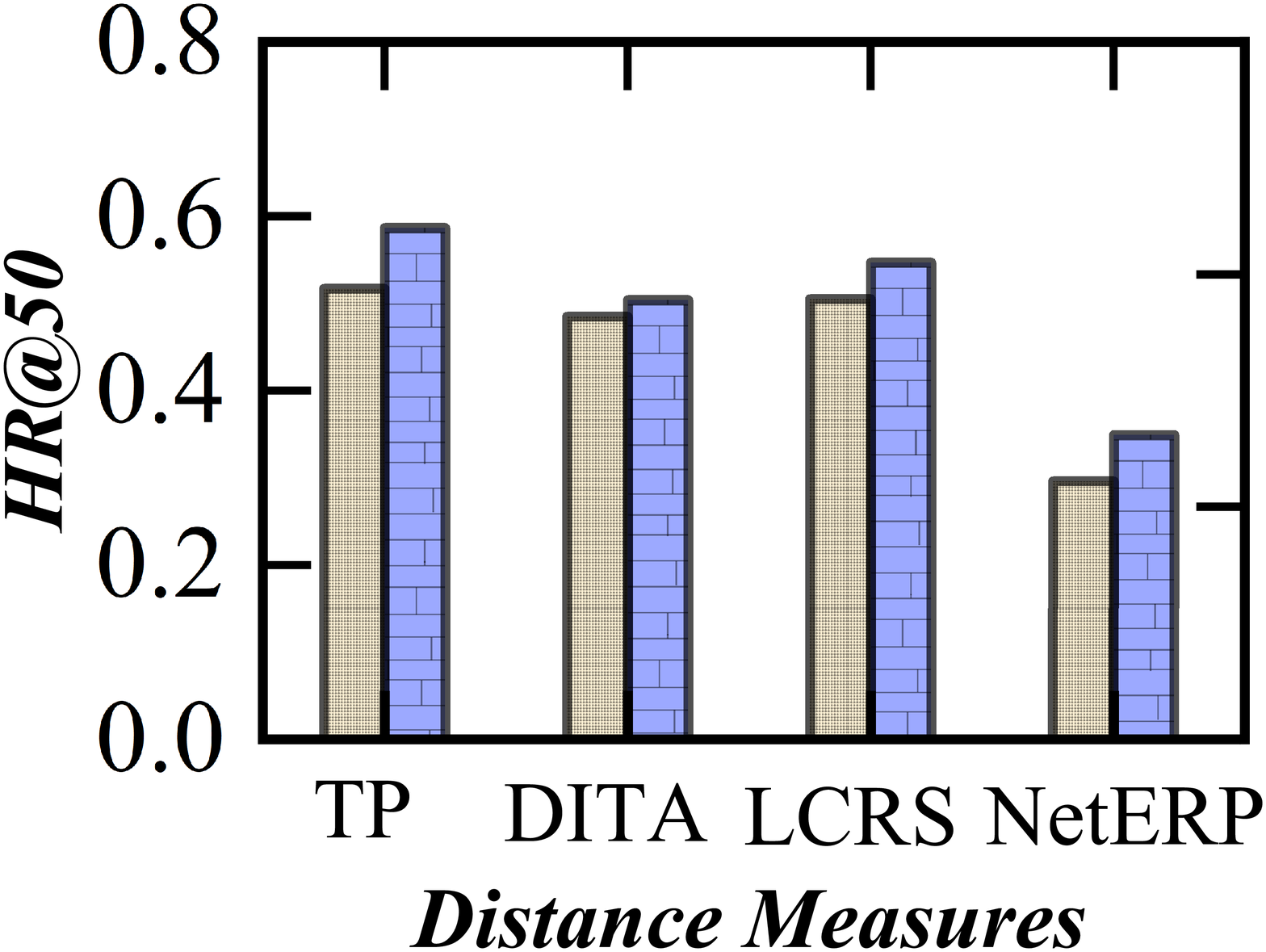}}
	\hspace{-0.25mm}
	\subfigure[Rome]{
		\includegraphics[width=0.23\textwidth]{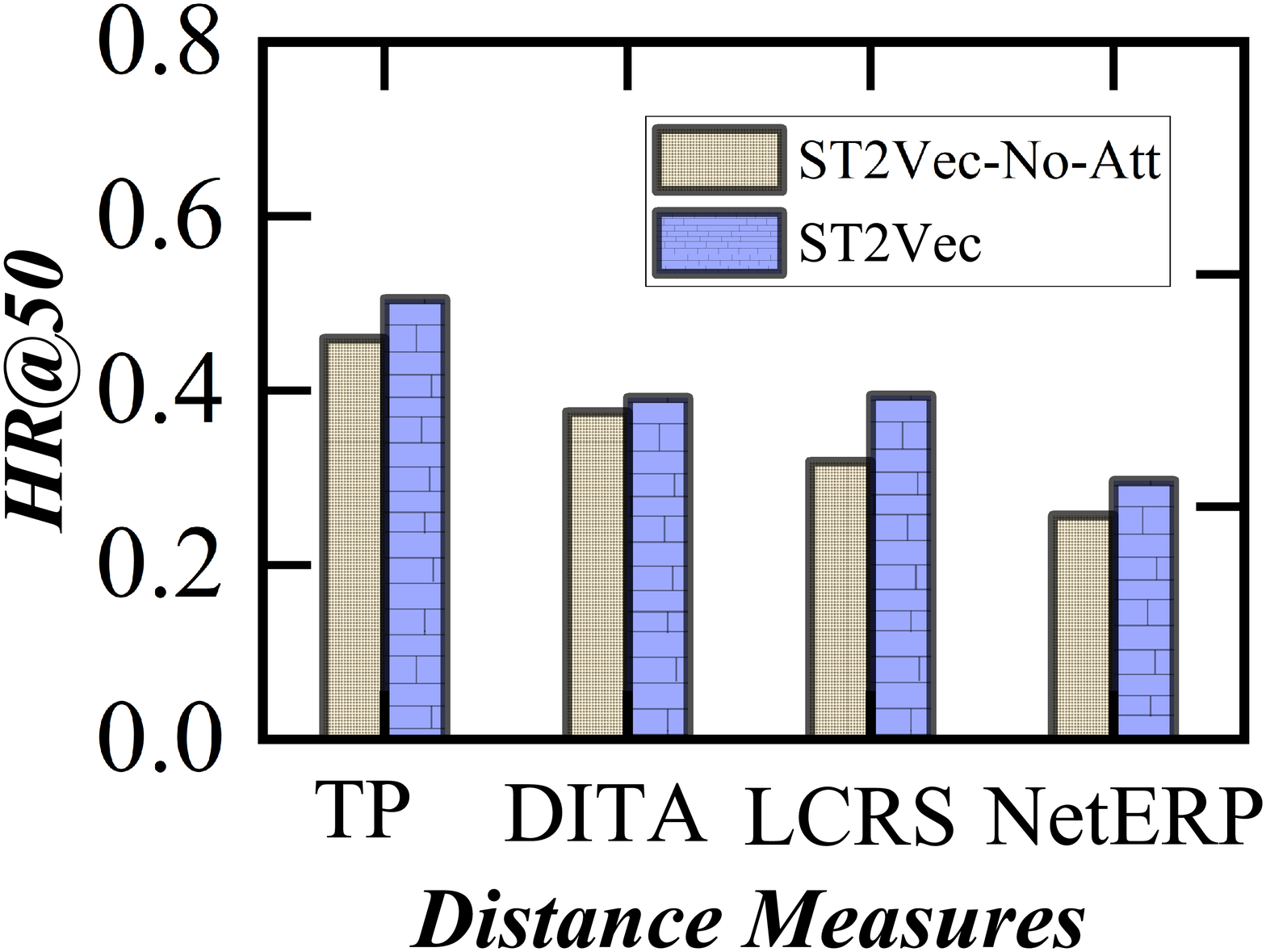}}
	\vspace{-5mm}
	\caption{ST2Vec Performance vs. with/without Attention}
	\vspace{-2mm}
	\label{fig:ModelAnalysisofattention}
\end{figure}

\begin{figure}[tb]
	\centering
	\subfigure[T-Drive]{
		\includegraphics[width=0.23\textwidth]{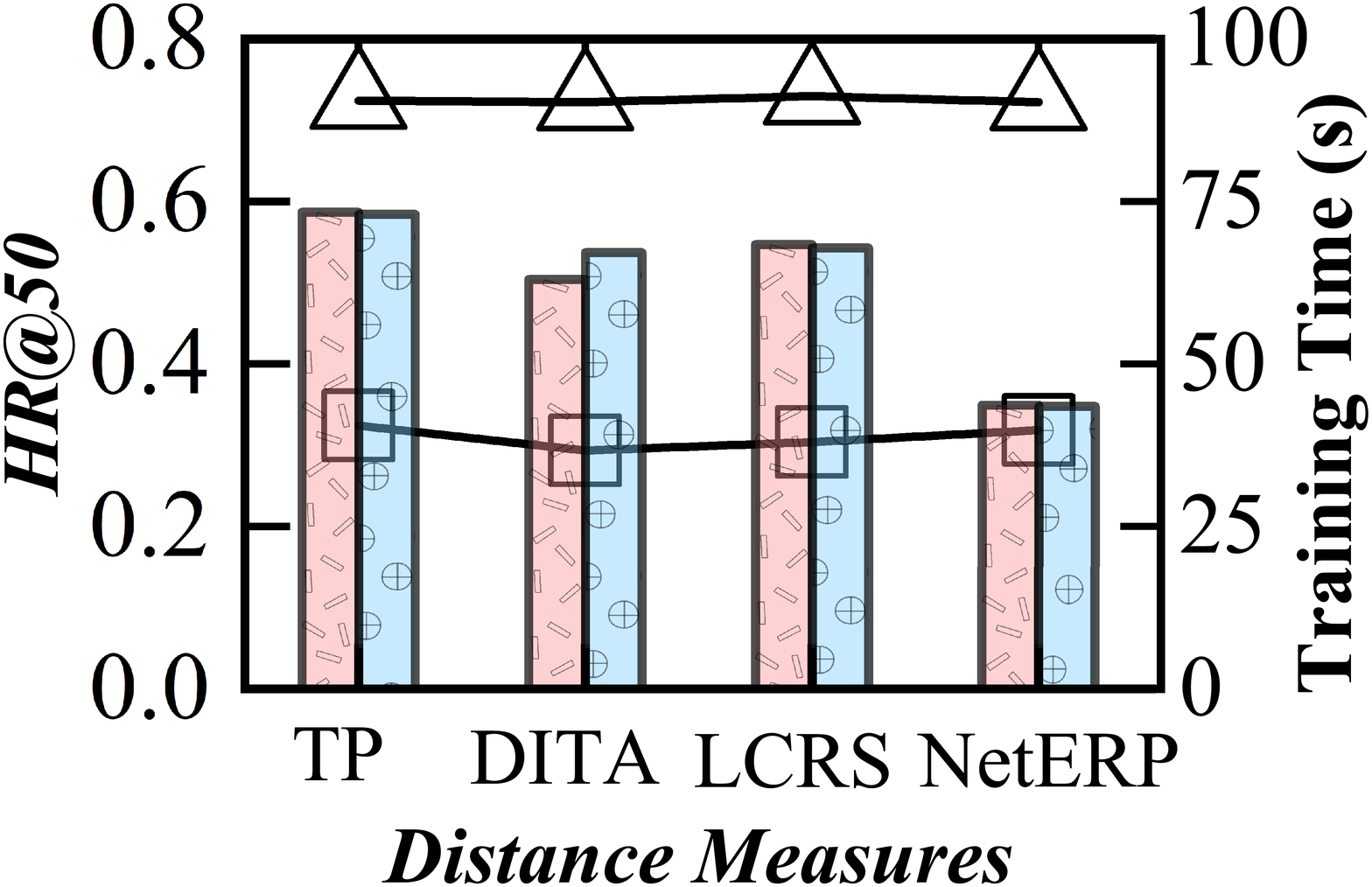}}
	\hspace{-0.25mm}
	\subfigure[Rome]{
		\includegraphics[width=0.23\textwidth]{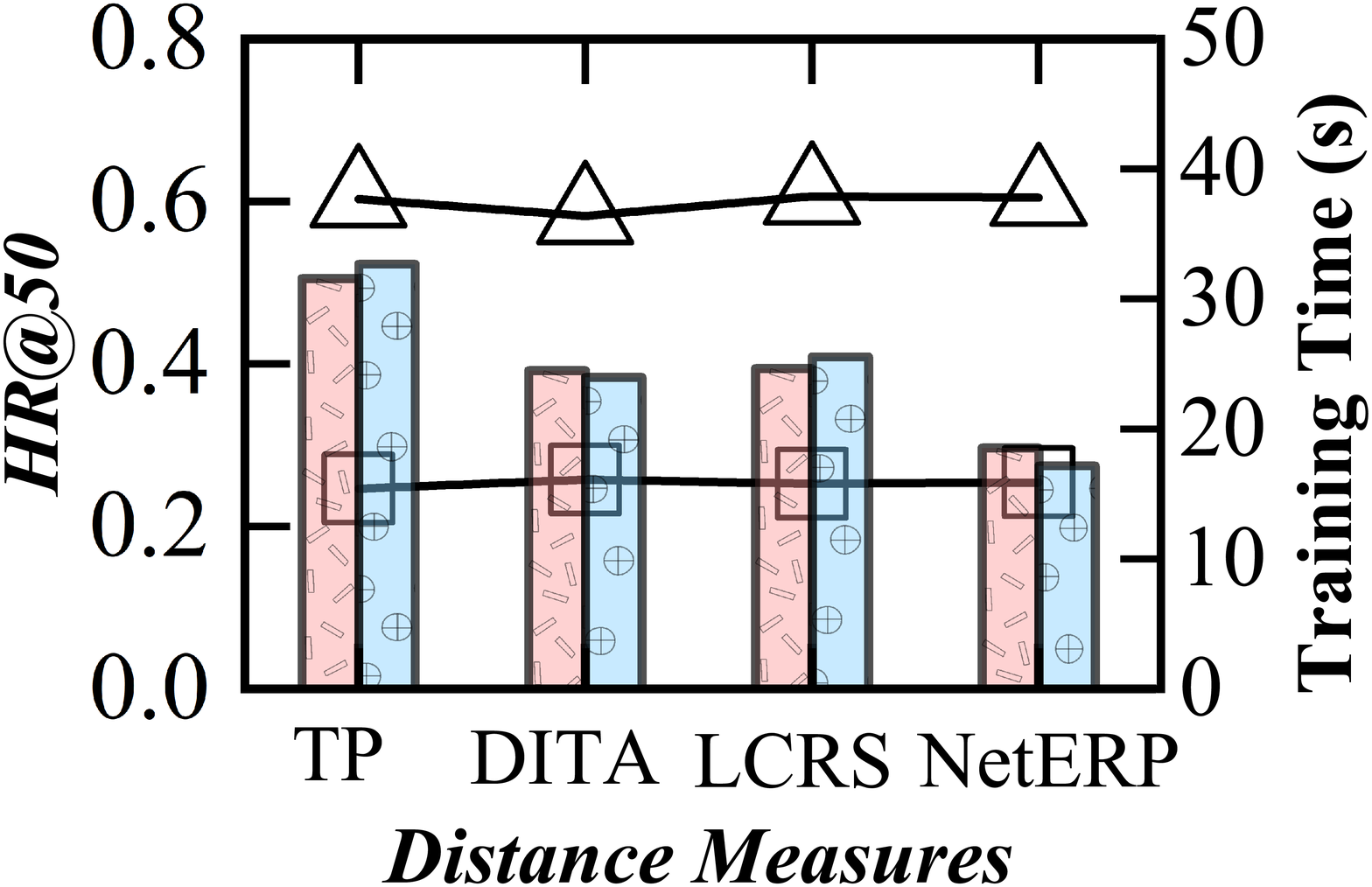}}
	\vspace{-5mm}
	\caption{ST2Vec Performance vs. Fusion Manners}
	\label{fig:ModelAnalysisoffusion}
	\vspace{-3mm}
\end{figure}

\begin{figure}[tb]
	\centering
	\subfigure[T-Drive]{
		\includegraphics[width=0.23\textwidth]{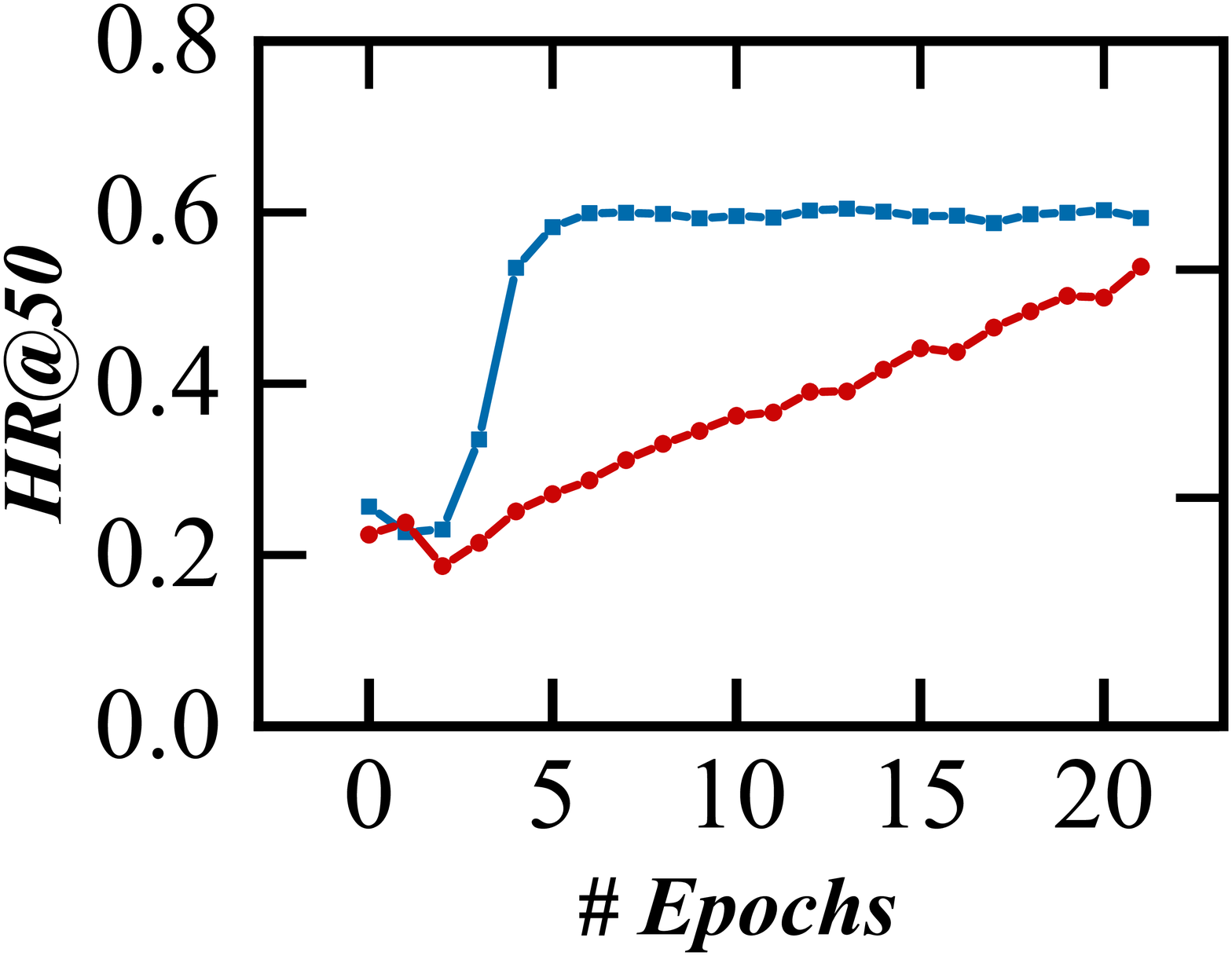}}
	\hspace{-0.25mm}
	\subfigure[Rome]{
		\includegraphics[width=0.23\textwidth]{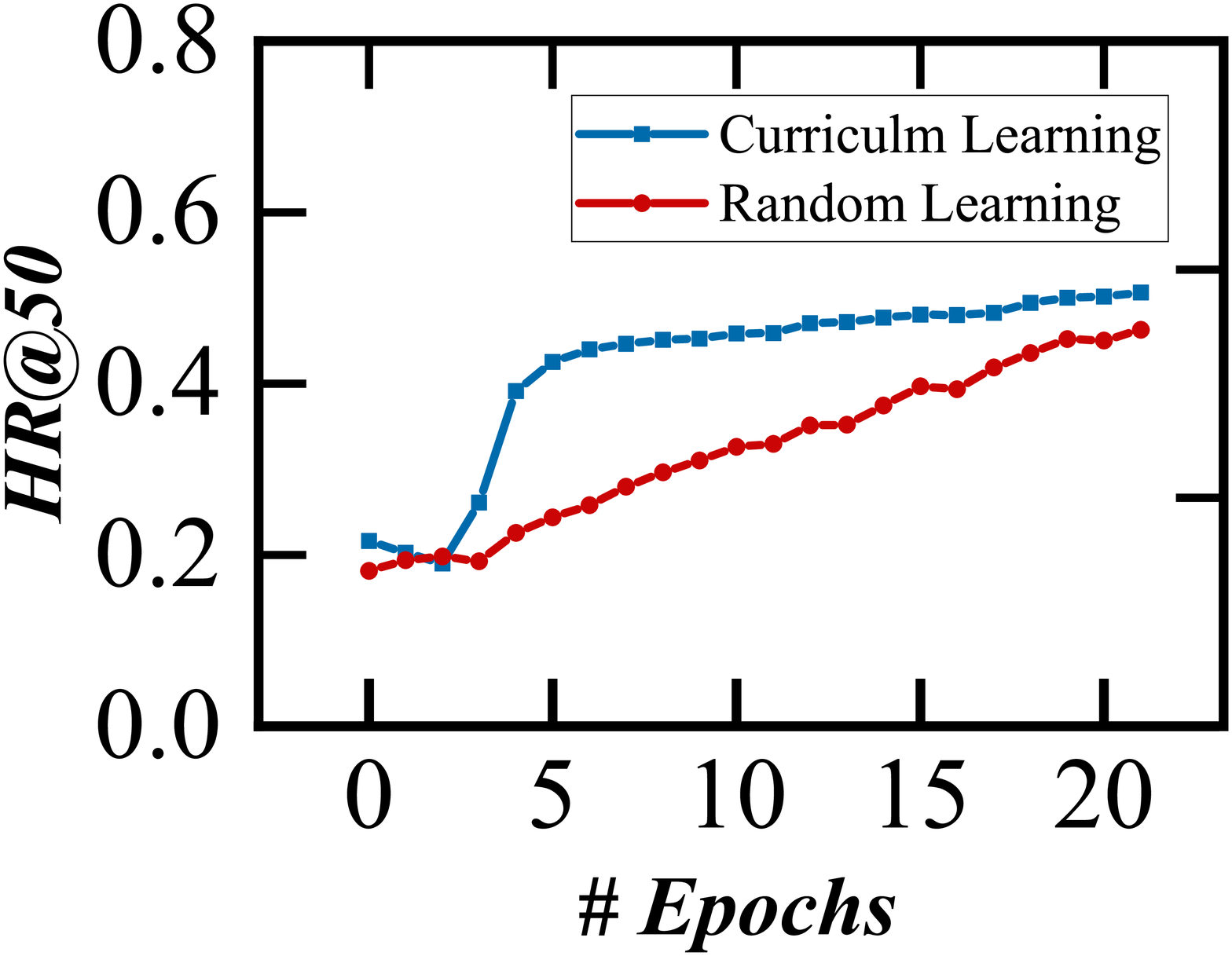}}
	\vspace{-5mm}
	\caption{The Convergence Curve of ST2Vec with respect to 20 epochs}
	\label{fig:ModelAnalysisofEpoch}
	\vspace{-3mm}
\end{figure}

\subsection{Ablation Study}
\noindent
\textbf{ST2Vec Performance vs. with/without Attention.} To study the effect of the attention mechanism on the performance, we remove it from ST2Vec and call the resulting model ST2Vec-No-Att. The HR@50 results on T-Drive are shown in Fig.~\ref{fig:ModelAnalysisofattention}, indicate that the spatial and temporal attention mechanisms are effective. Taking TP as an example, ST2Vec improves HR@50 over ST2Vec-No-Att from 0.51 to 0.58. 

\noindent
\textbf{ST2Vec Performance vs. Fusion Approach.} Second, to evaluate the effect of the fusion approach on model performance, we train ST2Vec using separate fusion (SF) and unified fusion (UF). Fig.~\ref{fig:ModelAnalysisoffusion} shows that ST2Vec using unified fusion achieves similar effectiveness to that using separate fusion. However, the ST2Vec-UF achieves fast model convergence than ST2Vec-SF, as SF features two separate LSM models that resulting in a double number of parameters to tune than UF. 

\noindent
\textbf{ST2Vec Performance vs. Curriculum/Random.} Finally, to evaluate the effect of curriculum learning on model performance, we consider all four distances. Fig.~\ref{fig:ModelAnalysisofEpoch} shows that the learning process guided by curriculum learning achieves faster convergence and higher computational quality (i.e., HR@50) than does random batch learning. This is because the curriculum strategy trains the model directionally by feeding training samples that vary from easy to hard (as discussed in Section~\ref{sec:method}-D).

\subsection{Efficiency Acceleration Study}
As a follow-up on the analysis in Section~\ref{sec:method}-E, we consider similarity computation using our ST2Vec-based method and a traditional pairwise based method. Table~\ref{table:time} reports the average time cost to perform top-50 spatio-temporal similarity search for each query trajectory with different data sizes, comparing the query efficiency under ST2Vec and Non-learning. As can be observed, ST2Vec achieves 200--400x speeds up over the non-learning based method.

\begin{figure*}[tb]
    \vspace{-3mm}
	\centering
	\includegraphics[width=1\textwidth]{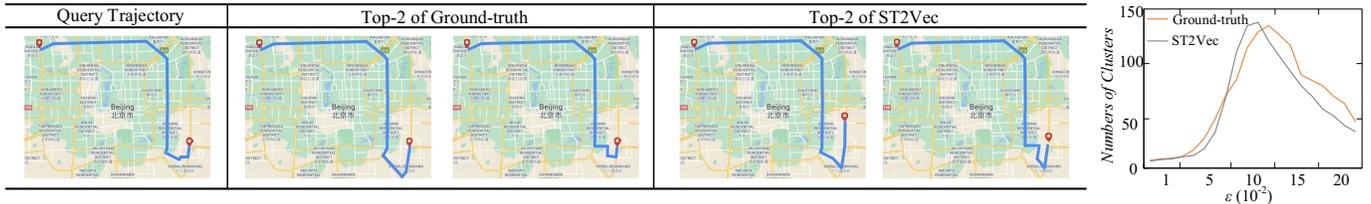}
	\vspace{-9mm}
	\caption{Case Studies: Top-$k$ Querying and Clustering}
	\label{fig:CaseStudies}
	\vspace{-3mm}
\end{figure*}

\subsection{Case Study}
We proceed to perform trajectory top-$k$ querying and clustering using T-Drive to examine the capabilities of ST2Vec intuitively. In terms of top-$k$ querying, we randomly choose one trajectory as the query trajectory. Then we plot its top-2 ground-truth trajectories according to TP as well as its top-2 similarity trajectories returned by ST2Vec. The left part of Fig.~\ref{fig:CaseStudies} plots different trajectories and shows that the trajectories returned by ST2Vec match the ground-truth trajectories very well. Next, we explore the effectiveness of ST2Vec using DBSCAN clustering when fixing the parameter \textit{minPts} at 10. Here, we also use TP. We compare the clustering results generated by the ground-truth and embedding based distances. As shown in the right part of Fig.~\ref{fig:CaseStudies}, the numbers of clusters in the two results share similar trends as $\epsilon$ grows, meaning that ST2Vec also works well for clustering analyses. 

\begin{table}[t]
\caption{Time Cost of Online Similarity Search on T-Drive}
\vspace{-2.5mm}
\begin{tabular}{p{1.2cm}<{\centering}|p{2cm}<{\centering}|p{0.7cm}<{\centering}p{0.7cm}<{\centering}p{0.7cm}<{\centering}p{0.7cm}<{\centering}}
\hline
Measures                & Methods     & 1k & 5k & 10k & 200k \\ \hline
\multirow{2}{*}{TP}     & Non-learning & 1.492s & 3.127s & 5.893s  & 117.832s   \\  
                        & \textbf{ST2Vec}      & 0.004s & 0.014s & 0.028s  & 0.521s   \\ \hline
\multirow{2}{*}{DITA}   & Non-learning & 0.921s & 3.301s & 6.291s  & 125.826s   \\  
                        & \textbf{ST2Vec}      & 0.004s & 0.015s & 0.028s  & 0.522s   \\ \hline
\multirow{2}{*}{LCRS}   & Non-learning & 1.292s & 4.614s & 8.784s  & 175.824s   \\  
                        & \textbf{ST2Vec}      & 0.004s & 0.014s & 0.028s  & 0.525s   \\ \hline
\multirow{2}{*}{NetERP} & Non-learning & 1.535s & 6.246s & 12.674s  & 253.481s   \\  
                        & \textbf{ST2Vec}      & 0.004s & 0.015s & 0.028s  & 0.522s   \\ \hline
\end{tabular}
\label{table:time}
\vspace{-5mm}
\end{table} 

%% file: Related.tex
\section{Related Work}
\label{sec:related}

We review related work on trajectory similarity in terms of non-learning-based methods and learning-based methods.

\noindent
\textbf{Non-learning-based methods}~\cite{XieLP17, shang2017trajectory, shang2018dita, wang2018torch, wang2019fast, Yuan019, KoideXI20} of trajectory similarity computation rely on well defined similarity measures and associated acceleration techniques. Here, we focus on popular similarity measures, while a comprehensive coverage of free-space based similarity measures is available elsewhere~\cite{SuLZZZ20, SousaBL20}. Network-aware similarity computation techniques first map original trajectories into road-network paths that consist of vertices or segments. Then, they define similarity measures based on classic distance measures such as Hausdorff~\cite{AtevMP10}, DTW~\cite{YiJF98}, LCSS~\cite{vlachos2002discovering}, and ERP~\cite{ChenN04}, generally by aggregating the distances between road vertices or segments of two trajectories. For example, Koide et al.~\cite{KoideXI20} propose NetERP by aggregating shortest path-distances between the vertices of two trajectories. Based on LCSS, Wang et al.~\cite{wang2018torch, wang2019fast} propose the Longest Overlapping Road Segments (LORS) for trajectory similarity computation and clustering. Similarly, Yuan et al.~\cite{Yuan019} propose the direction-aware Longest Common Road Segments (LCRS). These methods typically have quadratic computational complexity, as they rely on computations for aligned point pairs. 
Moreover, non-learning-based methods rely on hand-crafted heuristics, failing to exploit information hidden in trajectories. 

\noindent
\textbf{Learning-based methods}~\cite{deeprepresentation, seed, subsimilar, YangW0Q0021, HanWYS021} are becoming increasingly popular in recent years, as they feature the success of deep learning technologies, i.e., powerful approximation capability. The learning-based methods learn distance functions via neural networks that embed input trajectories and approximate given distance measures. This way, trajectory embeddings are generated that enable fast trajectory similarity computation and downstream analyses. Li et al.~\cite{deeprepresentation} propose t2vec that addresses the high computational cost of traditional methods while taking into account low sampling rates and the influence of noisy points. Nevertheless, t2vec was designed for trajectory representation learning, not similarity computation. Yao et al.~\cite{seed} propose NEUTRAJ, which employs metric learning to approximate trajectory similarity for different free-space based distance measures. Further, Zhang et al.~\cite{subsimilar} propose Traj2SimVec, which considers sub-trajectory similarity in the learning process. Zhang et al.~\cite{YangW0Q0021} propose T3S, which utilizes attention function to improve the performance. Despite the efforts of these studies, they all target spatial trajectory similarity in free space and cannot model the complex dependence of road networks. Recently, Han et al.~\cite{HanWYS021} develop GTS, which takes spatial trajectory similarity learning into road network context and achieves the state-of-the-art performance. Nevertheless, GTS is spatial-oriented similarity learning while ignoring the temporal aspect of trajectories. More specifically, GTS is designed for POI-based spatial trajectory similarity computation. For trajectories that share the same or neighbor POIs but with totally different traveling paths, GTS treats them as similar to each other. Besides, GTS only learns a single type of distance measure (i.e., TP~\cite{shang2017trajectory}, an extension of Hausdorff distance), while ST2Vec accommodates a series of popular measures including TP, DITA, LCRS, and NetERP.

%% file: Conclusion.tex
\section{Conclusions}
\label{sec:conclusion}

We propose ST2Vec, a representation learning based architecture for spatio-temporal similarity learning in road networks while enabling a range of trajectory measures. Extensive experiments using three real data sets confirm ST2Vec is capable of improved higher effectiveness, efficiency, and scalability than state-of-the-art methods. Also, similarity-based case studies of top-$k$ querying and clustering demonstrate the potential of ST2Vec for downstream analytics. In the future, it is of interest to integrate ST2Vec into spatial database management, thus enabling more types of trajectory analyses.